\documentclass[pdflatex,sn-basic]{sn-jnl}

\usepackage{graphicx}%
\usepackage{multirow}%
\usepackage{amsmath,amssymb,amsfonts}%
\usepackage{amsthm}%
\usepackage{mathrsfs}%
\usepackage[title]{appendix}%
\usepackage{xcolor}%
\usepackage{textcomp}%
\usepackage{manyfoot}%
\usepackage{booktabs}%
\usepackage[linesnumbered,ruled]{algorithm2e}
\usepackage{listings}%
\usepackage{dsfont}
\usepackage{pifont}
\usepackage[table]{xcolor}  
\usepackage{caption}
\SetKwInput{KwInput}{Input}
\SetKwInput{KwOutput}{Output}

\theoremstyle{thmstyleone}%
\theoremstyle{thmstyletwo}%

\theoremstyle{thmstylethree}%

\begin{document}

\title[Article Title]{PSScreen V2: Partially Supervised Multiple Retinal Disease Screening}


\author[1]{\fnm{Boyi} \sur{Zheng}}\email{boyi.zheng@oulu.fi} 

\author[2]{\fnm{Yalin} \sur{Zheng}}\email{yalin.zheng@liverpool.ac.uk}

\author[3]{\fnm{Hrvoje} \sur{Bogunovi\'c}}\email{hrvoje.bogunovic@meduniwien.ac.at}

\author*[1]{\fnm{Qing} \sur{Liu}}\email{qing.liu@oulu.fi}

\affil*[1]{\orgdiv{Center for Machine Vision and Signal Analysis (CMVS)}, \orgname{University of Oulu}, \orgaddress{\city{Oulu}, \postcode{90570}, \country{Finland}}}

\affil[2]{\orgdiv{Department of Eye and Vision Science}, \orgname{ University of Liverpool}, \orgaddress{\city{Liverpool}, \postcode{L7 8TX}, \country{U.K.}}}

\affil[3]{\orgdiv{Center for Medical Data Science}, \orgname{Medical University of Vienna}, \orgaddress{\city{Vienna}, \postcode{1090}, \country{Austria}}}


\abstract{In this work, we propose PSScreen V2, a partially supervised self-training framework for multiple retinal disease screening. Unlike previous methods that rely on fully labelled or single-domain datasets, PSScreen V2 is designed to learn from multiple partially labelled datasets with different distributions, addressing both label absence and domain shift challenges. To this end, PSScreen V2 adopts a three-branch architecture with one teacher and two student networks. The teacher branch generates pseudo labels from weakly augmented images to address missing labels, while the two student branches introduce novel feature augmentation strategies: Low-Frequency Dropout (LF-Dropout), which enhances domain robustness by randomly discarding domain-related low-frequency components, and Low-Frequency Uncertainty (LF-Uncert), which estimates uncertain domain variability via adversarially learned Gaussian perturbations of low-frequency statistics. Extensive experiments on multiple in-domain and out-of-domain fundus datasets demonstrate that PSScreen V2 achieves state-of-the-art performance and superior domain generalization ability. Furthermore, compatibility tests with diverse backbones, including the vision foundation model DINOv2, as well as evaluations on chest X-ray datasets, highlight the universality and adaptability of the proposed framework. The codes are available at \url{https://github.com/boyiZheng99/PSScreen_V2}.}



\maketitle

\section{Introduction}

\begin{figure*}[!t] 
    \centering 
    \includegraphics[width=1.0\textwidth]{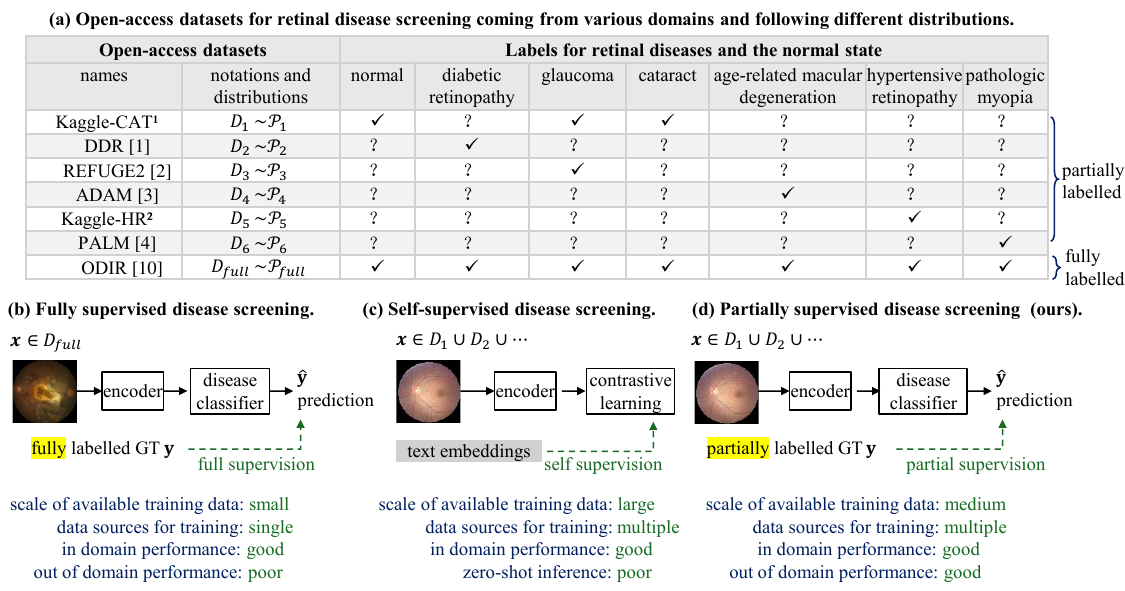} 
    \caption{Examples of open-access datasets for retinal disease screening and model comparisons under the three learning paradigms. (a) lists open-access datasets where ``\checkmark" indicates labels for diseases are available while ``?" denotes labels are not available. From (b) to (d), we illustrate the pipelines and characteristics of the fully supervised screening model usually trained with a fully labelled dataset, the self-supervised screening model trained with large-scale image-text pairs, and the partially supervised screening model trained with multiple partially labelled datasets.} 
    \label{fig:motivation} 
\end{figure*}

Automated early detection of retinal disease with fundus images is crucial for efficient and cost-effective large-scale population screening. Fueled by the release of open-access datasets summarized in Fig.~\ref{fig:motivation}(a) for specific retinal diseases such as Kaggle-CAT\footnote{\href{https://www.kaggle.com/datasets/jr2ngb/cataractdataset}{Kaggle-CAT}\label{cat}} for normal condition, glaucoma and cataract, DDR \citep{DDR_2019} for diabetic retinopathy, REFUGE2 \citep{REFUGE2_2022_arXiv} for glaucoma, ADAM \citep{ADAM_2022_TMI} for age-related macular degeneration, PALM \citep{PALM_2024} for myopia screening, Kaggle-HR\footnote{\href{https://www.kaggle.com/datasets/harshwardhanfartale/hypertension-and-hypertensive-retinopathy-dataset}{Kaggle-HR}\label{HR}} for hypertensive retinopathy etc., numerous works \citep {LAT_2021_CVPR,CA-Net_2023_TIM,SVPL_2024_TMI,C2x-FNet_2025_TIM} have been developed to train disease-specific screening models on individual training datasets. Although these models are promising in screening for specific diseases on images within specific domains, they are still far from real-world applications where screening for as many retinal diseases as possible on images from various or even unseen domains is desired. Developing a screening model for multiple retinal diseases with strong domain generalization ability is of great significance, yet remains challenging.

The first option is to train a disease screening model in a fully supervised way with fully labelled training data as illustrated in Fig.~\ref{fig:motivation}(b). For example, TrustDetector \citep{TrustDetector_2024_MICCAI} trains a screening model for three retinal diseases as the dataset collected only provides labels for three diseases while the method~\citep{ODIR_2021} trains the model on ODIR \citep{ODIR_2021} for multiple retinal disease screening. Although they achieve promising performances, the scale of training samples is small and they assume that training and test images share the same distribution, which limits their generalization ability to out-of-domain data. To increase the scale of training samples, \cite{Retina-1M_2024_TMI} collect and manually annotate a large-scale dataset e.g. Retina-1M to enable the fully supervised training. However, the annotation is labour-intensive and costly. 

The second option is to leverage the adaptability of foundation models such as FLAIR \citep{FLAIR_2025_MIA} and RET-CLIP \citep{RET-CLIP_2024_MICCAI} which are trained with large-scale image-text pairs in a self-supervised learning way and to detect multiple retinal disease via zero-shot inference, as illustrated in Fig.~\ref{fig:motivation}(c). Although the need for large-scale fully annotated training data is bypassed, the performance for screening is usually poor \citep{challenge_2024_MIA} due to lack of training with labelled samples.

The third option is to fully utilize existing partially labelled datasets and train a multiple retinal disease screening model in a partially supervised learning way as shown in Fig. \ref{fig:motivation}(d). On one hand, the need for full annotations for all interest of diseases is bypassed. On the other hand, the available annotations for partial diseases are fully leveraged for learning, thereby achieving better performances than zero-shot inference. However, this approach faces two challenges: 1) label absence for partial diseases in datasets specialized for specific diseases; and 2) domain shift among datasets from various medical sites caused by differences in imaging protocols, devices, and patient demographics, etc.

To address the two challenges, we propose \textbf{PSScreen V2}, a self-training \textbf{P}artially \textbf{S}upervised framework for multiple retinal disease \textbf{Screen}ing. It is an upgraded version of our previous work PSScreen V1~\citep{PSScreen_2025_BMVC} and inherits the core ideas of self-training paradigm to address the label absence issue and text-guided semantic decoupling to guide the screening model focus on disease-specific semantic regions. To address the challenge of domain shift, we propose two novel frequency-domain feature augmentations: low-frequency dropout (\textbf{LF-Dropout}) and low-frequency uncertainty (\textbf{LF-Uncert}). LF-Dropout augments features by randomly dropping low-frequency components while preserving the high-frequency components. LF-uncertainty augments feature by introducing Gaussian noises to randomly perturb the low-frequency components. These are motivated by the prior knowledge that low-frequency components are generally related to low-level statistics e.g., brightness and colour distribution, which are often domain-related, whilst high-frequency components are often related to edges and boundaries of structures such as tiny lesions and vessels etc. in retinal images, which are mostly disease-related and domain-invariant~\citep{FDA_2020_CVPR, ALOFT_2023_CVPR}. Perturbing low-frequency components enforces the network learn domain invariant features while maintaining the high-frequency components encourages the network learn the intrinsic, disease-related features for screening.

We validate the effectiveness of our \textbf{PSScreen V2} on six in-domain and six out-of-domain open-access fundus datasets. The results demonstrate that it outperforms eight state-of-the-art methods in screening performance as well as generalization ability to out-of-domains. We also explore the compatibility of our method with various backbones and the vision foundation model DINOv2~\citep{DINOv2_2024_TMLR}. The results show that, with our \textbf{PSScreen V2}, screening performances have been significantly improved. Furthermore, we evaluate \textbf{PSScreen V2} on three chest X-ray datasets, which demonstrates that \textbf{PSScreen V2} can also be applied to other disease classification problems, highlighting the versatility of \textbf{PSScreen V2}. We summarize our contributions as follows:
\begin{itemize}
    \item We propose \textbf{PSScreen V2}, a novel learning paradigm, partially supervised self-training framework which trains a single model from multiple partially labelled datasets for multi-disease screening.
    \item We propose \textbf{LF-Dropout}, which augments features by randomly dropping the low-frequency components of features, enabling the model to adapt to diverse domains.
    \item We propose \textbf{LF-Uncert}, which adversarially learns the distributions of the statistics of low-frequency components to characterise the domain uncertainty and augments features by sampling statistics from the distributions, thereby improving the model's generalization ability to unseen domains.
    \item Extensive experiments on six in-domain and six out-of-domain fundus datasets demonstrate that \textbf{PSScreen V2} achieves state-of-the-art performance. 
    \item  PSScreen v2 demonstrates strong compatibility with multiple backbone architectures and the vision foundation
    model DINOv2~\citep{DINOv2_2024_TMLR}, consistently improving screening performance over the baseline. Moreover, evaluations on chest X-ray datasets for thoracic diseases classification further validate the versatility of \textbf{PSScreen V2} in medical imaging.
\end{itemize}

This work is an extension of our previous paper PSScreen \citep{PSScreen_2025_BMVC} published in BMVC 2025. The new contributions include: (1) a self-training partially supervised learning framework with three branches for multiple retinal disease screening; (2) two novel feature augmentation strategies, i.e., LF-Dropout and LF-Uncert which significantly enhance the model’s domain generalization capability; (3) Extensive experiments validate the effectiveness of the proposed framework, including evaluations on multiple backbones and the vision foundation model DINOv2 under various parameter-efficient fine-tuning strategies, additional evaluations on chest X-ray datasets, and explanation analysis using concept activation vectors.

\section{Related Work}
\subsection{Multiple Retinal Disease Screening} 
\textbf{Fully Supervised Methods.} These methods typically rely on fully annotated datasets for model training. For example, \cite{TrustDetector_2024_MICCAI} propose TrustDetector for three retinal diseases. \cite{RAG_2024_MICCAI} use cross-attention to model correlations between global features and disease conditions, enabling the disentanglement of global features into disease-specific representations for multi-retinal disease screening. However, retinal abnormalities are small and scattered, and aforementioned methods fail to explicitly model the relationships between local abnormalities and overall disease diagnosis, limiting their ability to capture subtle lesions. To address this limitation, \cite{SIGraph_2025_AAAI} develop a GNN-based method, SIGraph, to capture lesion spatial distribution patterns. \cite{HVT_2025_AAAI} propose a forward attention mechanism which models relationships between inter-level features to ensure that subtle lesion details captured by lower layers are enhanced by the contextual understanding provided by higher layers. While these methods achieve promising performance, they are generally trained on single-source datasets with limited diversity. Thus, they often suffer from performance degradation when applied to out-of-domain samples. Building large-scale fully annotated datasets (e.g., Retina-1M~\citep{Retina-1M_2024_TMI}) could alleviate this issue, but annotation is expensive and labor-intensive.

\textbf{Zero-shot Inference with Foundation Models.} 
Recently, retinal foundation models such as RETFound~\citep{RETFound_2023_Nature}, FLAIR~\citep{FLAIR_2025_MIA}, and RET-CLIP~\citep{RET-CLIP_2024_MICCAI} have attracted 
considerable attention for their outstanding ability to learn general representations from large-scale unlabelled retinal images or image-text pairs, and excel in adapting to diverse downstream tasks. It is intuitive to employ zero-shot inference with those retinal foundation models for multiple retinal disease screening without any annotation costs. However, compared to full supervised task-specific screening models, the screening performances remain unsatisfying~\citep{FLAIR_2025_MIA}.

\subsection{Partially Supervised Learning for Image Classification}
\textbf{Partially Supervised Learning for Natural Image Classification.} Partially supervised learning tackles the problem arising when only partial class labels are available for each image. The naive solution is to assume that the unlabelled classes are negative, which leads to false negatives. To mitigate the adverse effects by the ``negative assumption", BoostLU~\citep{BoostLU_2023_CVPR} scales up the discriminative regions in class activation map (CAM) so that they resemble those obtained under full annotation while IGNORE~\citep{IGNORE_ECCV_2024} identifies and rejects the false negatives.  \cite{GRLoss_IJCAI_2024} propose to employ a soft pseudo-labelling mechanism to compensate the missing labels and design a generalized robust loss to cope with the noise introduced by pseudo labels.
Differently, SST~\citep{SST_2022_AAAI} and HST~\citep{HST_2024_IJCV} exploit the semantic co-occurrence and transfer knowledge of labelled classes while CALDNR~\citep{CALDNR_2024_TMM} utilizes the semantic similarity to generate pseudo labels for unlabelled classes. In PGCL~\citep{prompt_NN_2025}, CLIP \citep{CLIP_2021_ICML} is employed to guide the class-level representation decoupling, thereby reducing the visual confusion and yielding more reliable pseudo labels. Instead, SARB~\citep{SARB_2022_AAAI} complements unknown labels from images with known labels by blending representations. All these methods are dedicated to solving the label absent issue while ignoring the domain shift issue which limits their applications in medical fields.

\textbf{Partially Supervised Learning for Disease Classification.} In medical imaging, the challenge of partial class annotation is more severe and common due to the high cost and expertise required for annotation, yet few studies explore partially supervised learning. As pioneers, \cite{MPC_2024_TMI} propose MPC, the first partially supervised chest X-ray image classification method, which adopts a weak-to-strong consistency framework to correct the potential positive labels which are incorrectly treated as negatives in the most naive solution. FSP~\citep{FSP_TMI_2025}, the first partially supervised retinal disease screening method, trains two networks to generate pseudo labels for each other through curriculum co-pseudo labeling and active sample selection to recover the complete annotation information. Concurrently, we propose PSScreen V1~\citep{PSScreen_2025_BMVC} which is the first model for multi-disease screening trained with multiple datasets from diverse domains. PSScreen V1 \citep{PSScreen_2025_BMVC} adopts a weak-to-strong self-training framework to address label absence and performs uncertainty-based feature augmentation to enhance domain generalization against domain shift. Different to MPC \citep{MPC_2024_TMI} and FSP \citep{FSP_TMI_2025} which overlook the domain shift issue, our \textbf{PSScreen V2} inherits the core ideas of PSScreen V1 \citep{PSScreen_2025_BMVC} and addresses the two issues simultaneously.

\section{Proposed Method}
In this section, we provide a detailed description of the proposed \textbf{PSScreen V2}.

\subsection{Problem Formulation and Framework Overview} 
\textbf{Formulation.} Suppose that (1) a meta-dataset $\mathcal{D}=\left\{D_1,D_2,\dots,D_K\right\}$ consists of $K$ partially labelled datasets, collected from various medical sites and following different distributions, (2)  $(\mathbf{x}, \mathbf{y})$ denotes a sample from $\mathcal{D}$ where $\mathbf{x}$ is an image and $\mathbf{y}\in \{1,0,-1\}^T$ is the labels for $T$ retinal diseases. For each disease $t$, $y_t=1/0$ indicates positive/negative and $y_t=-1$ indicates an unknown label. To simplify, we use an indicator vector $\boldsymbol{\delta}=\mathds{1}_{\{1,0\}}(\mathbf{y})$ to indicate whether the label for each disease is known or unknown where $\mathds{1}(\cdot)$ is the indicator function.  Our goal is to train a model with $\mathcal{D}$ to predict the probabilities of $T$ diseases and to generalize well to diverse and even unseen domains.

\begin{figure*}[!t] 
    \centering 
    \includegraphics[width=1\textwidth]{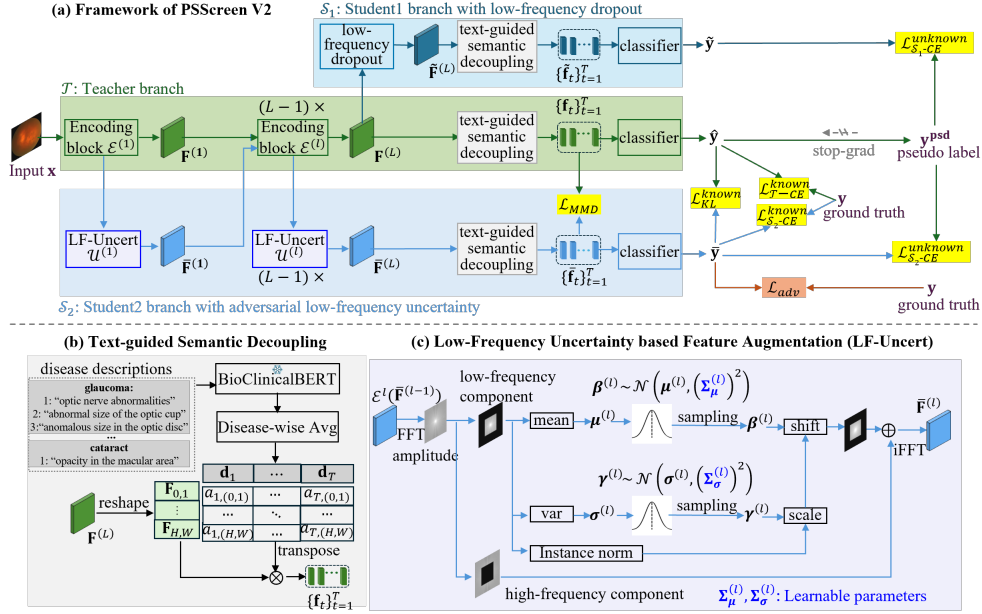} 
    \caption{(a) Illustration of the PSScreen V2. PSScreen V2 adopts a three-branch self-training framework, where two student branches are supervised by pseudo labels from a common $\mathsf{Teacher}$ $\mathcal{T}$. $\mathsf{Student1}$ $\mathcal{S}_{1}$ augments features via low-frequency dropout (LF-Dropout), while $\mathsf{Student2}$ $\mathcal{S}_{2}$ augments features via low-frequency uncertainty (LF-Uncert). All three branches share a backbone, a text-guided semantic decoupling module and a multi-label classifier. (b) An illustration of text-guided semantic decoupling module. (c) An illustration of LF-Uncert.}
    \label{fig:framework_psscreen_v2} 
\end{figure*}

\textbf{Overview.} As shown in Fig.~\ref{fig:framework_psscreen_v2}(a), to solve the partially supervised learning problem with multiple datasets from diverse domains, our \textbf{PSScreen V2} adopts a three-branch self-training framework. $\mathsf{Teacher}$  $\mathcal{T}$ takes the weakly augmented images as input and generates pseudo labels to supervise the learning of the two student branches. $\mathsf{Student1}$ $\mathcal{S}_1$ applies low-frequency dropout to augment features and encourages the screening model to be robust to diverse domains. $\mathsf{Student2}$ $\mathcal{S}_2$ applies low-frequency uncertainty based feature augmentation to model the domain uncertainty, thereby enhancing the model's generalization to unseen domains. In what follows, we will elaborate on the details of each branch.

\subsection{$\mathsf{Teacher}$ $\mathcal{T}$ for Pseudo Label Generation}

$\mathcal{T}$ consists of (1) an image encoder with $L$ encoding blocks $\{\mathcal{E}^{(l)}\}_{l=1}^{L}$, which takes the weakly augmented image as input and maps it to feature maps $\{\mathbf{F}^{(l)}\}_{l=1}^{L}$, and (2) a text-guided semantic decoupling module which decouples the feature maps by the encoder to disease-aware feature vectors, and (3) a multi-label classifier which predicts the probabilities of diseases and generates pseudo labels for unknown classes. 

\textbf{Text-Guided Semantic Decoupling.}
To focus on disease-related semantic regions, inspired by ~\citep{semantic_decoup_2019_ICCV, prompt_NN_2025}, we propose a text-guided semantic decoupling module. As shown in Fig.~\ref{fig:framework_psscreen_v2}(b), we use BioClinicalBERT \citep{BioClinicalBERT_2019} to encode the textual expert knowledge about each disease~\citep{FLAIR_2025_MIA}, then average the text embeddings of each disease to obtain the disease-wise text embedding $\{\mathbf{d}_t\}_{t=1}^T$. For the image feature maps $\mathbf{F}^{(L)} \in \mathbb{R}^{C \times H \times W }$ where $C$ is the number of channels and $H$ and $W$ are the spatial dimensions, extracted by the image encoder, following~\citep{semantic_decoup_2019_ICCV}, we compute the attention score $\alpha_{t, (h,w)}$ between each feature $\mathbf{F}_{h,w}$ at location $(h,w)$ and the disease-wise text embedding $\mathbf{d}_t$ via:
\begin{equation}
\alpha_{t, (h,w)}  = \frac{\text{exp} \left( \mathbf{v}^\top \tanh\left( \mathbf{W}_1 \mathbf{F}^{(L)}_{h,w} \odot \mathbf{W}_2 \mathbf{d}_t  \right)\right)}{\sum_{i,j} \text{exp} \left(  \left( \mathbf{v}^\top \tanh\left( \mathbf{W}_1 \mathbf{F}^{(L)}_{i,j} \odot \mathbf{W}_2 \mathbf{d}_t \right) \right)\right)} \;,
\label{eq:att}
\end{equation}
where $\odot$ denotes the the Hadamard product, and $\mathbf{W}_1$, $\mathbf{W}_2$, and $\mathbf{v}$ are learnable weights. Finally, we obtain the text-guided disease-aware feature vectors via aggregation:
\begin{equation}
\mathbf{f}_{t} =\sum_{h,w} \alpha_{t, (h,w)} \cdot \mathbf{F}^{(L)}_{h,w}\;.
\end{equation}

\textbf{Multi-retinal Disease Classification.} With the disease-aware feature vectors $\{\mathbf{f}_{t}\}_{t=1}^{T}$, we use multi-label classifier to predict the risks of diseases and the partial Binary Cross-Entropy (partial-BCE) loss~\citep{partial_loss_2019_CVPR} is imposed to guide the learning:
\begin{align}
\mathcal{L}^{known}_{\mathcal{T}\text{-}CE}(\mathbf{y},\hat{\mathbf{y}}) 
= - \frac{1}{\|\boldsymbol{\delta}\|_1}
\sum_{t=1}^T\delta_t\cdot \left({y}_t\cdot\log(\hat{{y}}_t) + (1-{y}_t)\cdot\log(1 - \hat{{y}}_t)\right),
\label{eq:l-ce}
\end{align}
where $\hat{\mathbf{y}}$ is the prediction and the loss is only computed for known classes. 

\textbf{Pseudo Label Generation.} To supervise the learning of the two student branches, we simply use thresholding to generate pseudo labels for unknown classes. Specifically, for disease $t$ whose label is unknown, its pseudo label ${y}^{psd}_t$ is produced via:
\begin{equation}
{y}^{psd}_t = 
\begin{cases}
    1, & \text{if~} {\delta}_t=0 \text{~and~} \hat{{y}}_t > \tau\\
    0, & \text{if~} {\delta}_t=0 \text{~and~} \hat{{y}}_t < 1- \tau \\
    -1 &\text{else}
\end{cases}
\label{eq:psd}
\end{equation}
where $\tau$ is a threshold and set to 0.95 in experiments. For simplicity, we use a vector $\boldsymbol{\zeta} = \mathds{1}_{\{1,0\}}(\mathbf{y}^{psd})$ to indicate whether a pseudo label is assigned to an unknown class.

\subsection{$\mathsf{Student1}~\mathcal{S}_1$ with Low-Frequency Dropout}
To improve generalization across diverse domains, we design a \textbf{L}ow-\textbf{F}requency \textbf{Dropout} (\textbf{LF-Dropout}) for feature augmentation. Our key insight is that low-frequency components are generally associated with domain-specific statistics such as brightness and color distributions, whereas high-frequency components are typically related to edges and boundaries of subtle lesions and vessels, which are usually indicative of disease~\citep{FDA_2020_CVPR, ALOFT_2023_CVPR}. Based on this, we propose LF-Dropout to randomly drop low-frequency components, simulating diverse domain variations while preserving high-frequency components. As shown in Fig. \ref{fig:framework_psscreen_v2}(a), in $\mathcal{S}_1$, LF-Dropout is applied to features $\mathbf{F}^{(L)}$, followed by a text-guided semantic decoupling module and a multi-label classifier to produce the predictions for consistency learning. 

\textbf{Low-frequency Dropout.} Given the feature maps $\mathbf{F}^{(L)}$ by the image encoder, we first apply 2D Fast Fourier Transform (2D-FFT)~\citep{FFT_1982} independently to each channel and transform them to the frequency domain:
\begin{equation}
\left(\boldsymbol{\mathcal{A}}, \boldsymbol{\mathcal{P}} \right) = \text{2D-FFT}(\mathbf{F}^{(L)})\;,
\label{eq:fft}
\end{equation}
where $\boldsymbol{\mathcal{A}}$ and $\boldsymbol{\mathcal{P}}$ denote the amplitude and phase of frequency components, respectively. We shift $\boldsymbol{\mathcal{A}}$ so that the low-frequency components are centered in the spectrum. We then randomly set the magnitudes of low-frequency components within a centered $r\times r$ region to zero with probability $p$, while keeping the remaining unchanged. Since this operation is analogous to dropout, we term it as Low-Frequency-Dropout (LF-Dropout). Formally, it can be expressed as:  
\begin{equation}
    \text{LF-Dropout}(\boldsymbol{\mathcal{A}}, p, r) = \mathbf{M}\odot \boldsymbol{\mathcal{A}}\;,
    \label{Eq:LF-Dropout}
\end{equation}
where $\mathbf{M}$ is a mask whose elements are set by:
\begin{equation}
    \mathbf{M}_{i,j}\sim
    \begin{cases}
        \text{Bernoulli}(1-p) & \text{if } (i,j)\in \mathcal{R}_r \;  \\
        \text{Bernoulli}(1) & \text{otherwise}\;;
    \end{cases}
\end{equation}
where $\text{Bernoulli}(\cdot)$ denotes the Bernoulli distribution, and $\mathcal{R}_r$ is the set of indices 
corresponding to the centered low-frequency region:
\begin{equation}
\mathcal{R}_r = \left\{ (i,j) \;\middle|\;
\left| i - \left\lfloor \tfrac{H}{2} \right\rfloor \right| \le \tfrac{r}{2}, \;
\left| j - \left\lfloor \tfrac{W}{2} \right\rfloor \right| \le \tfrac{r}{2}
\right\}.
\end{equation}
Finally, we use the inverse FFT (iFFT) to transform the features back to spatial domain and obtain the augmented features:
\begin{equation}
\tilde{\mathbf{F}}^{(L)} = \text{2D-iFFT}\left(\text{LF-Dropout}(\boldsymbol{\mathcal{A}}, p, r), {\boldsymbol{\mathcal{P}}}\right).
\end{equation}

\textbf{Pseudo Label Supervision.} 
 To guide learning for unknown classes, we use the pseudo labels produced by the $\mathsf{Teacher}$ branch and impose partial-BCE loss:
\begin{align}
\mathcal{L}^{\text{unknown}}_{\mathcal{S}_1\text{-}CE}
(\mathbf{y}^{psd}, \tilde{\mathbf{y}})
= - \frac{1}{\|\boldsymbol{\zeta}\|_1}
\sum_{t=1}^T \zeta_{t} \
\Big(
y_t^{psd}\,\log(\tilde{y}_t)
+ (1-y_t^{psd})\,\log(1-\tilde{y}_t)
\Big),
\label{eq:psl_AD}
\end{align}
where $\tilde{\mathbf{y}}$ is the predictions by $\mathcal{S}_1$ and $\boldsymbol{\zeta}$ indicates whether the pseudo label is assigned to the disease, and $\mathbf{y}^{psd}$ is the pseudo labels by Eq. (\ref{eq:psd}). The supervision with pseudo labels, in turn, helps the model learn to adapt to diverse domains.

\subsection{$\mathsf{Student2}~\mathcal{S}_2$ with Adversarial Low-Frequency Uncertainty}
To improve generalization to unseen domains, we design \textbf{L}ow-\textbf{F}requency \textbf{Uncert}ainty based feature augmentation (\textbf{LF-Uncert}). Our insight is that domain characteristics are often relative to lighting conditions and imaging systems etc., and characterized by low-frequency statistics of images. Naturally, we propose LF-Uncert which uses the uncertain Gaussian distributions to characterize the low-frequency uncertainty, thereby simulating the unpredictable domain variability in frequency domain. With LF-Uncert, features are augmented by resampling the low-frequency statistics and reconstruction. To ensure the augmented space is as broad as possible, adversarial learning is used to learn the parameters of the Gaussian distributions. 

\textbf{Low-Frequency Uncertainty based Augmentation.}
As shown in Fig.~\ref{fig:framework_psscreen_v2}(c), in $\mathcal{S}_2$, LF-Uncert module is attached to each encoding block of the image encoder. Each takes $\mathcal{E}^{(l)}\left(\bar{\mathbf{F}}^{(l-1)}\right)$ as input, and generates the augmented features $\bar{\mathbf{F}}^{(l)}$, where $\bar{\mathbf{F}}^{(l)}={\mathbf{F}}^{(l)}$ when $l=1$. Specifically, in module  $\mathcal{U}^{(l)}$, the 2D FFT is first applied to transform $\mathcal{E}^{(l)}\left(\bar{\mathbf{F}}^{(l-1)}\right)$ into frequency domain with Eq. (\ref{eq:fft}), and the amplitude is denoted as $\boldsymbol{\bar{\mathcal{A}}}$. For the low-frequency components $\boldsymbol{\bar{\mathcal{A}}}^{LF}$ within the centered $r\times r$ region, we compute their channel-wise statistics, i.e., mean and standard deviation via:
\begin{equation}
\boldsymbol{\mu}^{(l)} = \frac{1}{r^2} \sum_{h=1}^{r} \sum_{w=1}^{r} \boldsymbol{\bar{\mathcal{A}}}^{LF}_{h,w}\;,
\label{Eq:LF-Uncert-mean}
\end{equation}
and
\begin{equation}
\boldsymbol{\sigma}^{(l)} = \sqrt{\frac{1}{r^2} \sum_{h=1}^{r} \sum_{w=1}^{r}\left(\boldsymbol{\bar{\mathcal{A}}}^{LF}_{h,w}-\boldsymbol{\mu}^{(l)} \right)^2}\;.
\label{Eq:LF-Uncert-var}
\end{equation}
Then, we perturb the statistics via adding noises randomly:
\begin{equation}
\boldsymbol{\beta}^{(l)} = \boldsymbol{\mu}^{(l)} + \boldsymbol{\epsilon_\mu}^{(l)},\quad \boldsymbol{\gamma}^{(l)} = \boldsymbol{\sigma}^{(l)} + \boldsymbol{\epsilon_\sigma}^{(l)},
\end{equation}
where
\begin{equation}
\boldsymbol{\epsilon_\mu}^{(l)}\sim \mathcal{N}\left(\bold{0},  (\boldsymbol{\Sigma}_{\boldsymbol{\mu}}^{(l)})^2\right),\quad \boldsymbol{\epsilon_\sigma}^{(l)}\sim \mathcal{N}\left(\bold{0},  (\boldsymbol{\Sigma}_{\boldsymbol{\sigma}}^{(l)})^2\right).
\label{eq:noise_dist}
\end{equation}
Thereafter, following FSC \citep{FSC_2025_TIP}, we apply AdaIN \citep{AdaIN_ICCV_2017} to the low-frequency components and obtain the augmented low-frequency components:
\begin{equation}
 \boldsymbol{\bar{\mathcal{A}}}^{LF}_{aug} = \boldsymbol{\gamma}^{(l)} \left( \frac{\boldsymbol{\bar{\mathcal{A}}}^{LF} - \boldsymbol{\mu^{(l)}}}{\boldsymbol{\sigma^{(l)}}} \right) + \boldsymbol{\beta}^{(l)}.
\end{equation}
Finally, we apply inverse 2D FFT to transform the features from frequency domain to spatial domain and obtain the augmented features $\bar{\mathbf{F}}^{(l)}$.

\textbf{Multi-retinal Disease Classification.} With the augmented features $\bar{\mathbf{F}}^{(l)}$, same as in the teacher branch, the text-guided semantic decoupling module is attached to yield disease-aware feature vectors $\{\mathbf{\bar{f}}_t\}_{t=1}^{T}$ and classifier is attached to obtain the predictions $\bar{\mathbf{y}}$. We use both the ground truth labels $\mathbf{y}$ and the pseudo labels $\mathbf{y}^{psd}$ to supervise the learning: 
\begin{equation}
\mathcal{L}_{\mathcal{S}_2\text{-}CE}(\mathbf{y},\mathbf{y}^{psd}, \bar{\mathbf{y}}) =  \mathcal{L}^{known}_{\mathcal{S}_2\text{-}CE}(\mathbf{y},\bar{\mathbf{y}}) + \lambda_1  \mathcal{L}^{\text{unknown}}_{\mathcal{S}_2\text{-}CE}
(\mathbf{y}^{psd}, \bar{\mathbf{y}})\;,
\label{Eq:Loss-S2-CE}
\end{equation}
where 
\begin{align}
\mathcal{L}^{known}_{\mathcal{S}_2\text{-}CE}(\mathbf{y},\bar{\mathbf{y}}) 
= \frac{1}{\|\boldsymbol{\delta}\|_1}
\sum_{t=1}^T\delta_t\cdot \left({y}_t\cdot\log(\bar{{y}}_t) + (1-{y}_t)\cdot\log(1 - \bar{{y}}_t)\right)\;,
\label{eq:L_S2_CE_known}
\end{align}
and
\begin{align}
\mathcal{L}^{\text{unknown}}_{\mathcal{S}_2\text{-}CE}
(\mathbf{y}^{psd}, \bar{\mathbf{y}})
= - \frac{1}{\|\boldsymbol{\zeta}\|_1}
\sum_{t=1}^T \zeta_{t} \
\Big(
y_t^{psd}\,\log(\bar{y}_t)
+ (1-y_t^{psd})\,\log(1-\bar{y}_t)
\Big)\;.
\label{eq:L_S2_CE_unknown}
\end{align}

\textbf{Consistency Learning Between $\mathcal{T}$ and $\mathcal{S}_2$.} To transfer the knowledge from teacher branch $\mathcal{T}$ to student branch $\mathcal{S}_2$ and gradually improve $\mathcal{S}_2$, same as PSScreen V1 \citep{PSScreen_2025_BMVC}, feature-level and output-level consistency are imposed. At feature level, we enforce the consistency between disease-wise features by $\mathcal{S}_2$ and $\mathcal{T}$ via minimizing the MMD loss \citep{MMD_2015_ICML}:
\begin{equation}
\mathcal{L}_{MMD}(\mathbf{f}_{t},\bar{\mathbf{f}}_{t}) = \frac{1}{T} \sum_{t=1}^T \left\lVert\phi(\mathbf{f}_{t}) - \phi(\bar{\mathbf{f}}_{t})\right\rVert^2\;,
\label{eq:MMD_loss}
\end{equation}
where $\phi(\cdot)$ is the Gaussian kernel function in a reproducing kernel Hilbert space. At output level, we enforce the consistency between the predictions of the known classes via minimizing the KL divergence:
\begin{equation}
\mathcal{L}_{KL}^{known}(\hat{\mathbf{y}},\bar{\mathbf{y}}) = \frac{1}{\|\boldsymbol{\delta}\|_1} \sum_{t=1}^T \delta_t \cdot (\hat{\mathbf{y}}_t \cdot \log \frac{\hat{\mathbf{y}}_t}{\bar{\mathbf{y}}_t})\;.
\label{eq:sd_label_loss}
\end{equation}
Up to now, the total loss in $\mathcal{S}_2$ can be expressed as:
\begin{align}
    \mathcal{L}_{\mathcal{S}_2} = \mathcal{L}_{\mathcal{S}_2\text{-}CE}(\mathbf{y},\mathbf{y}^{psd}, \bar{\mathbf{y}}) + \lambda_2\mathcal{L}_{MMD} + \lambda_3\mathcal{L}_{KL}^{known}\;.
\label{eq:overall_loss_S2} 
\end{align}
\textbf{Adversarial LF-Uncert.} To augment the features as broad as possible, unlike ALOFT \citep{ALOFT_2023_CVPR} which estimates the noise variances in Eq. (\ref{eq:noise_dist}) empirically from the batch samples, we learn $\{(\boldsymbol{\Sigma}_{\boldsymbol{\mu}}^{(l)})^2\}_{l=1}^{L}$ and $\{(\boldsymbol{\Sigma}_{\boldsymbol{\sigma}}^{(l)})^2\}_{l=1}^{L}$ adversarially. Specifically, the student branch $\mathcal{S}_2$ is guided to learn the variances of noise such that the statistics of the augmented features differ greatly from those of the input features, causing the predictions based on the augmented features to deviate maximally from the ground truth labels. To achieve this, we maximize the KL divergence between the prediction $\mathbf{\bar{y}}$ by $\mathcal{S}_2$ and the ground truth $\mathbf{y}$:
\begin{equation}
    \arg\max_{\{(\boldsymbol{\Sigma}_{\boldsymbol{\mu}}^{(l)})^2\}_{l=1}^{L}, \{(\boldsymbol{\Sigma}_{\boldsymbol{\sigma}}^{(l)})^2\}_{l=1}^{L} } KL(\mathbf{y}||\mathbf{\bar{y}})\;,
\end{equation}
which is equivalent to minimize the adversarial loss below:
\begin{equation}
\mathcal{L}_{adv}(\mathbf{y},\bar{\mathbf{y}}) = -\frac{1}{\|\boldsymbol{\delta}\|_1} \sum_{t=1}^T \delta_t \cdot ({\mathbf{y}}_t \cdot \log \frac{{\mathbf{y}}_t}{{\bar{\mathbf{y}}}_t})\;. 
\label{eq:L-adv}
\end{equation}

\subsection{Model Training}
In total, the overall loss, excluding the adversarial loss $\mathcal{L}_{adv}$, can be expressed as:
\begin{equation}    \mathcal{L}_{total}=\mathcal{L}_{\mathcal{T}-CE}^{known} + \mathcal{L}_{\mathcal{S}_1-CE}^{unknown} + \mathcal{L}_{\mathcal{S}_2}\;.
\label{eq:L-total}
\end{equation}
We adversarially train our PSScreen V2 by iteratively minimizing the $\mathcal{L}_{total}$ to update parameters excluding $\{(\boldsymbol{\Sigma}_{\boldsymbol{\mu}}^{(l)})^2,(\boldsymbol{\Sigma}_{\boldsymbol{\sigma}}^{(l)})^2\}_{l=1}^{L}$, and minimizing $\mathcal{L}_{adv}$ to update  $\{(\boldsymbol{\Sigma}_{\boldsymbol{\mu}}^{(l)})^2,(\boldsymbol{\Sigma}_{\boldsymbol{\sigma}}^{(l)})^2\}_{l=1}^{L}$. The details are illustrated in Algorithm~\ref{alg:psscreenv2}. During the inference, we only use the $\mathsf{Teacher}$ branch for multi-disease detection.

\begin{algorithm}[t!]
\caption{Adversarial training for PSScreen V2}
\label{alg:psscreenv2}
\KwInput{ \\
\hspace*{2em} Training samples in meta data $\mathcal{D}=\{(\mathbf{x}, \mathbf{y})\}$.\\
\hspace*{2em} Initialised PSScreen V2 with parameters: \\
\hspace*{4em} $\theta_{\mathcal{E}}$ in the image encoder,\\
\hspace*{4em} $\theta_{txt}=\{\mathbf{W}_1$, $\mathbf{W}_2$, $\mathbf{v}\}$ in Eq. (\ref{eq:att}),\\
\hspace*{4em} $\theta_{cls}$ in classifier,\\
\hspace*{4em} $\{(\boldsymbol{\Sigma}_{\boldsymbol{\mu}}^{(l)})^2,(\boldsymbol{\Sigma}_{\boldsymbol{\sigma}}^{(l)})^2\}_{l=1}^{L}$ in Eq. (\ref{eq:noise_dist}) in LF-Uncert.} 
\KwOutput{Optimized PSScreen V2.}
\For{Number of training iterations}{
    \tcp{\textcolor{gray}{Fix $\{(\boldsymbol{\Sigma}_{\boldsymbol{\mu}}^{(l)})^2,(\boldsymbol{\Sigma}_{\boldsymbol{\sigma}}^{(l)})^2\}_{l=1}^{L} $, update $\theta_{\mathcal{E}}, \theta_{txt}$ and $\theta_{cls}$.}}
    
    Forward $\textbf{x}$ and obtain $\mathbf{\hat{y}}, \mathbf{\tilde{y}}, \mathbf{\bar{y}}$ by $\mathcal{T}, \mathcal{S}_1, \mathcal{S}_2$, respectively.\;
    
    Generate pseudo labels $\mathbf{y}^{psd}$ using Eq. (\ref{eq:psd})\;
    
    Compute  $\mathcal{L}_{total}$ using Eq. (\ref{eq:L-total}).\;
    
    Update $\theta_{\mathcal{E}}, \theta_{text}$ and $\theta_{cls}$ by descending $\nabla\mathcal{L}_{total}$.\;
    
    \tcp{\textcolor{gray}{Fix $\theta_{\mathcal{E}}, \theta_{txt}, \theta_{cls}$, update $\{(\boldsymbol{\Sigma}_{\boldsymbol{\mu}}^{(l)})^2,(\boldsymbol{\Sigma}_{\boldsymbol{\sigma}}^{(l)})^2\}_{l=1}^{L} $.}}
    
    Compute the adversarial loss $\mathcal{L}_{adv}$ using Eq.~(\ref{eq:L-adv}).\;
    
    Update $\{(\boldsymbol{\Sigma}_{\boldsymbol{\mu}}^{(l)})^2,(\boldsymbol{\Sigma}_{\boldsymbol{\sigma}}^{(l)})^2\}_{l=1}^{L}$ by descending $\nabla\mathcal{L}_{adv}$.\; 
}
\Return{PSScreen V2}
\end{algorithm}

\section{Experiments on Retinal Disease Screening}
In this section, we first introduce the datasets and implementation details, then conduct extensive comparisons and ablation studies to validate the effectiveness, domain generalization capability, and compatibility of our \textbf{PSScreen V2}. Finally, an in-depth analysis with explainability tools is conducted.

\subsection{Experimental Setup}
\textbf{Datasets.} We follow PSScreen V1 \citep{PSScreen_2025_BMVC} and construct two combined datasets: (1) \textbf{Meta}~dataset, constituting of six datasets, i.e., Kaggle-CAT\footref{cat}, DDR~\citep{DDR_2019},  REFUGE2~\citep{REFUGE2_2022_arXiv}, ADAM~\citep{ADAM_2022_TMI}, Kaggle-HR\footref{HR}, and PALM~\citep{PALM_2024}, and (2) \textbf{Unseen}~dataset constituting of four datasets i.e., APTOS2019\footnote{\href{https://www.kaggle.com/competitions/aptos2019-blindness-detection/data}{APTOS2019}\label{aptos}}, ORIGA$^{\text{light}}$~\citep{ORIGA_2010}, HPMI\footnote{\href{https://figshare.com/articles/dataset/HPMI_A_retinal_fundus_image_dataset_for_identification_of_high_and_pathological_myopia_based_on_deep_learning/24800232?file=49305304}{HPMI dataset on Figshare} \label{hpmi}}, and RFMiD~\citep{RFMiD_2021}. Meta is partially labelled for the normal state, diabetic retinopathy (DR), glaucoma, cataract, age-related macular degeneration (AMD), hypertensive retinopathy (HR), and pathologic myopia (PM). Among them,  DR is labelled with five grades and the other conditions with binary labels. For consistent formulation, we formulate the five-class DR grading problem as five binary classification tasks, aligning it with the binary setup used for other diseases and the normal condition. We use Meta for both training and in-domain
validation, and Unseen for out-of-domain validation. Tab.~\ref{tab:dataset_info} shows more details about Meta and Unseen. Additionally, we use \textbf{ODIR} \citep{ODIR_2021}, a fully labelled dataset, for out-of-domain validation and the preprocessing is same as that in PSScreen V1 \citep{PSScreen_2025_BMVC}. To further assess the model’s domain generalization ability under zero-shot inference, we follow \citep{FLAIR_2025_MIA} and validate on \textbf{ODIR200$\times$3}, a subset of 600 images that contains three types of labels, i.e., normal state, cataract, and pathologic myopia and with 200 images in each category. 

\begin{table*}[!t]
    \centering
    \caption{Details about datasets used. `N' is normal, `D' is diabetic retinopathy (DR), `G' is glaucoma, `C' is cataract, `A' is age-related macular degeneration (AMD), `H' is hypertensive retinopathy (HR), and `P' is pathologic myopia (PM).}
    \resizebox{1.0\textwidth}{!}{
    \begin{tabular}{cllcccccccccccc}
    \toprule
        Group & Dataset & Resolution & \#images & \multicolumn{3}{c}{Original Splitting} & \multicolumn{7}{c}{Labels} & train/test \\
        \cmidrule(lr){5-7} \cmidrule(lr){8-14}
        & & & & Train & Valid & Test & N & D & G & C & A & H & P & \\
        \midrule
        \multirow{6}{*}{Meta} & Kaggle-CAT & 2592 × 1728 or 2464×1632 & 600 & 360 & 120 & 120 & \checkmark &  & \checkmark & \checkmark &  &  &  & \checkmark/\checkmark \\
        & DDR & max: 5184×3456 min: 512×512 & 12522 & 6261 & 2504 & 3757 &  & \checkmark &  &  &  &  &  & \checkmark/\checkmark \\
        & REFUGE2 & max: 2124×2056 min: 1634×1634 & 2000 & 1200 & 400 & 400 &  &  & \checkmark &  &  &  &  & \checkmark/\checkmark \\
        & ADAM & 2124 × 2056 or 1444×1444 & 1200 & 400 & 400 & 400 &  &  &  &  & \checkmark &  &  & \checkmark/\checkmark \\
        & Kaggle-HR & 800×800 & 712 & 427 & 142 & 143 &  &  &  &  &  & \checkmark &  & \checkmark/\checkmark \\
        & PALM & 2124 × 2056 or 1444×1444 & 1200 & 400 & 400 & 400 &  &  &  &  &  &  & \checkmark & \checkmark/\checkmark \\
        \midrule
        \multirow{4}{*}{Unseen} & RFMiD & max: 4288×2848 min: 2048×1536 & 3200 & 1920 & 640 & 640 & \checkmark &  &  &  & \checkmark &  &  & \ding{55}/\checkmark \\
        & APTOS2019  & max: 4288×2848 min: 474×358 & 3662 & — & — & — &  & \checkmark &  &  &  &  &  & \ding{55}/\checkmark \\
        & ORIGA$^{\text{light}}$  & 3072 × 2048 & 650 & — & — & — &  &  & \checkmark &  &  &  &  & \ding{55}/\checkmark \\
        & HPMI & 512 × 512 & 4011 & — & — & — &  &  &  &  &  &  & \checkmark & \ding{55}/\checkmark \\
        \midrule
        ODIR & ODIR  & max: 5184×3456 min: 160x120 & 7949 & 6961 & — & 988 & \checkmark & \checkmark & \checkmark & \checkmark & \checkmark & \checkmark & \checkmark & \ding{55}/\checkmark \\
        ODIR 200×3 & ODIR 200x3 & max: 5184×3456 min: 868x793 & 600 & — & — & — & \checkmark &  &  & \checkmark &  &  & \checkmark & \ding{55}/\checkmark \\
        \botrule
    \end{tabular}}

    \label{tab:dataset_info}
\end{table*}

\textbf{Evaluation Metrics.} We follow \citep{ODIR_2021,SVPL_2024_TMI, PSScreen_2025_BMVC}, and use  
F-score and quadratic weighted kappa (QWK) as the evaluation metrics. Especially, the F‑score is computed with macro‑averaging, which treats each class equally and thus mitigates the impact of class imbalance. For the evaluation across multiple tasks and datasets, following PSScreen V1~\citep{PSScreen_2025_BMVC}, mean F-score ($mF$) and mean QWK ($mQWK$) are used.

\textbf{Implementation Details.} We follow PSScreen V1 \citep{PSScreen_2025_BMVC} and  crop the field of view from each fundus image, then pad the short side with zeros to equal length with the long side and resize it to 512×512. For image augmentation, we apply random scaling with a scale factor uniformly sampled from [0.8,1.2] with a probability of 0.5, followed by padding or cropping to maintain the input size. We then apply the augmentation strategies from~\citep{Multi_label_Transformer_2023_JBHI}, excluding Cutout. To prevent small-scale datasets in Meta from being underrepresented, an equal number of samples is drawn from each dataset in each batch during training.

We adopt ResNet-101~\citep{ResNet_2016_CVPR} pretrained on ImageNet~\citep{ImageNet_2009_CVPR} as the backbone and initialize other parameters randomly. The expert knowledge descriptions for the text-guided semantic decoupling module are borrowed from~\citep{FLAIR_2025_MIA}. Training uses the AdamW optimizer~\citep{AdamW_2017_ICLR} with a batch size of 16, a weight decay of 5×10$^{\text{-4}}$, and an initial learning rate of 1×10$^{\text{-5}}$ reduced by a factor of 10 every 10 epochs. We set  $r$ to 0.2 in Eq. (\ref{Eq:LF-Dropout}), Eq. (\ref{Eq:LF-Uncert-mean}) and Eq. (\ref{Eq:LF-Uncert-var}) and the drop probability $p$  in Eq. (\ref{Eq:LF-Dropout}) to 0.2. \textbf{PSScreen V2} is implemented with PyTorch and trained on two NVIDIA V100 GPUs with 32 GB RAM for 20 epochs.

\subsection{Comparison with Partially Supervised Learning Methods}
\textbf{Methods for Comparison.} We compare \textbf{PSScreen V2} with eight methods: MultiNets, MultiHeads, SST \citep{SST_2022_AAAI}, SARB \citep{SARB_2022_AAAI}, HST \citep{HST_2024_IJCV}, BoostLU \citep{BoostLU_2023_CVPR}, CALDNR \citep{CALDNR_2024_TMM}, and PSScreen V1~\citep{PSScreen_2025_BMVC}. Among them, MultiNets and MultiHeads are baselines, while the remaining ones are state-of-the-art methods for partially supervised learning. In detail, MultiNets trains multiple task-specific models and each responsible for a specific disease. MultiHeads consists of a single backbone network for feature extraction and multiple classification heads and each responsible for a specific disease classification. During training, losses are calculated only over labelled classes. For SST~\citep{SST_2022_AAAI}, SARB~\citep{SARB_2022_AAAI}, HST~\citep{HST_2024_IJCV}, BoostLU~\citep{BoostLU_2023_CVPR}, CALDNR~\citep{CALDNR_2024_TMM}, and PSScreen V1~\citep{PSScreen_2025_BMVC}, we follow the implementation details in original papers. For fair comparisons, all methods use ResNet-101 as the backbone.

\textbf{Results on Meta.} We separately report the F-score and QWK on Meta for \textbf{PSScreen V2} and the compared methods in Tab.~\ref{tab:f detail comparison in Meta} and Tab.~\ref{tab:qwk detail comparison in Meta}. Clearly, \textbf{PSScreen V2} achieves the best performance in terms of $mF$ and $mQWK$, with improvements of 0.3\% in $mF$ and 1.2\% in $mQWK$ over the second-best, i.e., PSScreen V1 \citep{PSScreen_2025_BMVC}. In detail, PSScreen V2 achieves the highest in F-score in classifying the normal state, glaucoma, and PM, and ranks second in screening HR among the seven classes. It is worth mentioning that the F-score for DR is substantially lower than those for others. This is because DR classification in this work is a five-level grading task where cases at level 1, i.e., mild DR, are rare and manifest as subtle microaneurysms, making them hard to detect and often confused with DR level 2, i.e., moderate DR. Consequently, the recall for level 1 is low, thereby lowering the average F-score across the five levels.

Among the compared methods, SST \citep{SST_2022_AAAI} and HST \citep{HST_2024_IJCV} utilize semantic co-occurrence for pseudo label generation, which are not always applicable in retinal disease screening, leading to suboptimal performance. BoostLU \citep{BoostLU_2023_CVPR} and CALDNR \citep{CALDNR_2024_TMM} treat all unknown classes as negatives during training, which introduces false negatives and hampers model learning. 

\begin{table*}[!t]
\centering
\caption{F-scores (\%) on Meta. The best and second-best are highlighted in bold and with an underline. Means and standard deviations are reported over three trials.}
\resizebox{1.0\textwidth}{!}{
\begin{tabular}{lcccccccccc}
\toprule
Methods             & T1: Normal & T2: DR & \multicolumn{3}{c}{T3: Glaucoma}                            & T4: Cataract & T5: AMD & T6: HR & T7: PM & $mF$ \\ 
\cmidrule{4-6}
                              & Kaggle-CAT       & DDR                      & REFUGE2                  & Kaggle-CAT          & Average & Kaggle-CAT          & ADAM               & Kaggle-HR & PALM              &                                    \\ 
\midrule
MultiNets                     & 78.1$_{\pm1.2}$  & \textbf{67.8$_{\pm0.9}$}   & 75.7$_{\pm1.3}$   & \textbf{89.5$_{\pm1.0}$}   & \underline{82.6$_{\pm1.0}$}   & 88.0$_{\pm2.0}$   & \underline{85.3$_{\pm1.7}$}   & 80.4$_{\pm3.7}$   & \textbf{96.7$_{\pm0.4}$}   & 82.7$_{\pm0.7}$          \\ 

MultiHeads                    & 86.8$_{\pm0.8}$   & 66.8$_{\pm1.0}$   & 77.2$_{\pm1.4}$   & 85.4$_{\pm0.4}$   & 81.3$_{\pm0.6}$   & 88.8$_{\pm3.2}$   & 84.4$_{\pm1.2}$   & 81.1$_{\pm1.7}$   & 96.3$_{\pm0.1}$   & 83.6$_{\pm0.7}$          \\  

SST$_{AAAI2022}$             & \underline{87.6$_{\pm0.8}$}   & \underline{67.1$_{\pm1.1}$}   & 73.6$_{\pm1.1}$   & 87.3$_{\pm1.5}$   & 80.5$_{\pm0.6}$   & 92.3$_{\pm2.8}$   & 83.7$_{\pm2.2}$   & 77.6$_{\pm3.4}$   & 95.2$_{\pm0.7}$   & 83.4$_{\pm1.4}$          \\ 

SARB$_{AAAI2022}$            & 86.2$_{\pm3.3}$ & 66.7$_{\pm1.6}$ & 77.0$_{\pm2.4}$ & 86.9$_{\pm0.7}$ & 81.9$_{\pm1.5}$ & 90.0$_{\pm0.9}$ & \textbf{85.9$_{\pm0.9}$} & 81.4$_{\pm2.1}$ & 96.0$_{\pm1.0}$ & 84.0$_{\pm0.9}$ \\

BoostLU$_{CVPR2023}$         & 80.2$_{\pm4.2}$   & 62.0$_{\pm1.3}$   & 72.6$_{\pm1.8}$   & 85.8$_{\pm3.1}$   & 78.6$_{\pm2.6}$   & 90.0$_{\pm2.0}$   & 76.8$_{\pm2.3}$   & 78.8$_{\pm1.5}$   & 95.9$_{\pm0.4}$   & 80.3$_{\pm0.5}$          \\ 

HST$_{IJCV2024}$              & 87.0$_{\pm0.5}$   & 67.0$_{\pm0.3}$   & 72.3$_{\pm2.5}$   & \underline{89.0$_{\pm3.1}$}   & 80.7$_{\pm0.8}$   & \underline{92.6$_{\pm1.7}$}   & 81.2$_{\pm2.1}$   & 79.6$_{\pm1.0}$   & 95.6$_{\pm1.0}$   & 83.4$_{\pm0.4}$          \\ 

CALDNR$_{TMM2024}$        & 82.5$_{\pm3.1}$   & 63.8$_{\pm1.3}$   & 74.6$_{\pm4.0}$   & 88.6$_{\pm1.6}$   & 81.6$_{\pm1.4}$   & \textbf{93.1$_{\pm1.5}$}   & 75.6$_{\pm7.3}$   & 81.8$_{\pm3.8}$   & 95.7$_{\pm0.2}$   & 82.0$_{\pm1.1}$          \\ 

PSScreen V1$_{BMVC2025}$    & 86.4$_{\pm1.7}$  & 64.7$_{\pm1.7}$  & \underline{77.7$_{\pm1.5}$} & \underline{89.0$_{\pm3.1}$}  & \textbf{83.3$_{\pm2.0}$}  & 89.1$_{\pm2.4}$  & 84.5$_{\pm1.4}$  & \textbf{85.0$_{\pm1.5}$}  & \underline{96.5$_{\pm0.4}$}  & \underline{84.2$_{\pm0.3}$}  \\ 

\textbf{PSScreen V2}      & \textbf{89.0$_{\pm1.7}$}  & 62.1$_{\pm1.4}$  & \textbf{78.2$_{\pm2.2}$} & 88.4$_{\pm0.4}$  & \textbf{83.3$_{\pm0.7}$}  & 92.3$_{\pm1.2}$  & 84.6$_{\pm0.9}$  & \underline{83.9$_{\pm2.2}$}  & \textbf{96.7$_{\pm0.5}$}  & \textbf{84.5$_{\pm0.5}$}  \\ 
\botrule
\end{tabular}}

\label{tab:f detail comparison in Meta}
\end{table*}

\begin{table*}[!t]
\centering
\caption{$QWK$ (\%) on Meta. The best and second-best are highlighted in bold and with an underline. Means and standard deviations are reported over three trials.} 
\resizebox{1.0\textwidth}{!}{
\begin{tabular}{lcccccccccc}
\toprule
Methods             & T1: Normal & T2: DR & \multicolumn{3}{c}{T3: Glaucoma}                            & T4: Cataract & T5: AMD & T6: HR & T7: PM & $mQWK$ \\ 
\cmidrule{4-6}
                              & Kaggle-CAT       & DDR                      & REFUGE2                  & Kaggle-CAT         & Average & Kaggle-CAT         & ADAM               & Kaggle-HR & PALM              &                                    \\ 
\midrule
MultiNets                     & 56.5$_{\pm2.6}$   & 88.1$_{\pm0.1}$   & 51.4$_{\pm2.6}$   & \textbf{79.0$_{\pm4.0}$}   & 65.2$_{\pm3.7}$   & 76.0$_{\pm4.1}$   & \underline{70.5$_{\pm3.5}$}   & 61.2$_{\pm7.4}$   & \textbf{93.5$_{\pm0.9}$}   & 73.0$_{\pm1.4}$          \\ 

MultiHeads                    & 73.6$_{\pm1.6}$   & 87.8$_{\pm0.8}$   & 54.5$_{\pm2.8}$   & 71.0$_{\pm0.7}$   & 62.7$_{\pm1.2}$   & 77.5$_{\pm6.5}$   & 68.7$_{\pm2.5}$   & 62.3$_{\pm3.5}$   & 92.6$_{\pm0.3}$   & 75.0$_{\pm1.3}$          \\ 

SST$_{AAAI2022}$            & \underline{75.2$_{\pm1.6}$}   & \underline{88.4$_{\pm0.2}$}   & 47.6$_{\pm2.1}$   & 74.6$_{\pm3.1}$   & 61.1$_{\pm1.4}$   & 84.7$_{\pm5.5}$   & 67.5$_{\pm4.2}$   & 55.2$_{\pm6.8}$   & 90.4$_{\pm1.3}$   & 74.7$_{\pm2.8}$          \\ 

SARB$_{AAAI2022}$           & 72.5$_{\pm6.7}$   & 87.3$_{\pm0.9}$   & 54.0$_{\pm4.7}$   & 77.1$_{\pm1.5}$   & 63.9$_{\pm3.0}$   & 80.0$_{\pm1.8}$   & \textbf{71.7$_{\pm1.7}$}   & 62.7$_{\pm4.3}$   & 92.0$_{\pm0.9}$   & 75.7$_{\pm1.8}$          \\ 

BoostLU$_{CVPR2023}$          & 60.8$_{\pm8.2}$   & 84.0$_{\pm2.2}$   & 46.1$_{\pm3.6}$   & 71.7$_{\pm6.0}$   & 57.6$_{\pm5.2}$   & 80.1$_{\pm4.0}$   & 54.7$_{\pm4.1}$   & 57.6$_{\pm3.0}$   & 91.8$_{\pm0.8}$   & 69.5$_{\pm1.1}$          \\ 

HST$_{IJCV2024}$            & 74.1$_{\pm0.9}$   & \textbf{88.8$_{\pm0.1}$}   & 45.0$_{\pm4.5}$   & 78.0$_{\pm6.2}$  & 61.5$_{\pm1.6}$   & \underline{85.3$_{\pm3.5}$}   & 62.5$_{\pm4.3}$   & 59.3$_{\pm1.8}$   & 91.3$_{\pm1.0}$   & 74.7$_{\pm0.8}$          \\ 

CALDNR$_{TMM2024}$          & 65.3$_{\pm5.9}$   & 87.0$_{\pm0.8}$   & 50.0$_{\pm7.3}$   & 77.2$_{\pm3.2}$   & 63.6$_{\pm2.4}$   & \textbf{86.3$_{\pm3.1}$}   & 52.4$_{\pm13.2}$  & 63.7$_{\pm7.5}$   & 91.5$_{\pm0.5}$   & 72.8$_{\pm2.2}$          \\ 

PSScreen V1$_{BMVC2025}$        & 73.0$_{\pm3.4}$  & 87.4$_{\pm0.2}$  & \underline{55.4$_{\pm3.0}$}  & \underline{78.1$_{\pm6.2}$}  & \textbf{66.8$_{\pm4.0}$}  & 78.3$_{\pm4.7}$  & 69.1$_{\pm2.7}$  & \textbf{70.1$_{\pm3.0}$}  & \underline{93.0$_{\pm0.9}$}  & \underline{76.8$_{\pm0.8}$}  \\ 

\textbf{PSScreen V2}                   & \textbf{78.0$_{\pm3.4}$}  & 86.3$_{\pm0.4}$  & \textbf{56.3$_{\pm4.4}$}  & 76.8$_{\pm1.8}$  & \underline{66.6$_{\pm1.4}$}  & 84.6$_{\pm2.5}$  & 69.2$_{\pm1.9}$  & \underline{67.8$_{\pm4.4}$}  & \textbf{93.5$_{\pm1.0}$}  & \textbf{78.0$_{\pm1.0}$}  \\ 

\botrule
\end{tabular}}
\label{tab:qwk detail comparison in Meta}
\end{table*}

\begin{table*}[t!]
    \centering
    \large
    \caption{F-score (\%) and QWK (\%) on Unseen. The best and second-best are highlighted in bold and with an underline. Means and standard deviations are reported over three trials.}
    \resizebox{1.0\textwidth}{!}{
    \begin{tabular}{lcccccccccccc}
        \toprule
        Methods & \multicolumn{2}{c}{T1: Normal} & \multicolumn{2}{c}{T2: DR} & \multicolumn{2}{c}{T3: Glaucoma} & \multicolumn{2}{c}{T4: AMD} & \multicolumn{2}{c}{T5: PM} & \multicolumn{2}{c}{Average} \\
        & \multicolumn{2}{c}{RFMID} & \multicolumn{2}{c}{APTOS} & \multicolumn{2}{c}{ORIGA} & \multicolumn{2}{c}{RFMID} & \multicolumn{2}{c}{HPMI} & \multicolumn{2}{c}{} \\
        \cmidrule(lr){2-3} \cmidrule(lr){4-5} \cmidrule(lr){6-7} \cmidrule(lr){8-9} \cmidrule(lr){10-11} \cmidrule(lr){12-13}
        & F-score & QWK & F-score & QWK & F-score & QWK & F-score & QWK & F-score & QWK & $mF$ & $mQWK$ \\
        \midrule
        MultiNets    & 68.1$_{\pm2.3}$ & 36.7$_{\pm4.2}$ & 41.3$_{\pm2.2}$ & 70.2$_{\pm5.8}$ & \textbf{71.3$_{\pm0.7}$} & \textbf{42.8$_{\pm1.5}$} & 48.3$_{\pm2.4}$ & 7.3$_{\pm1.1}$ & \textbf{83.1$_{\pm1.1}$} & \textbf{66.8$_{\pm2.1}$} & 62.4$_{\pm1.2}$ & 44.8$_{\pm2.3}$ \\
        MultiHeads   & 67.4$_{\pm4.1}$ & 36.3$_{\pm7.5}$ & 44.7$_{\pm2.7}$ & 77.7$_{\pm4.7}$ & 67.9$_{\pm1.8}$ & 36.7$_{\pm3.0}$ & 52.3$_{\pm2.0}$ & 9.9$_{\pm2.1}$ & 79.6$_{\pm0.4}$ & 60.4$_{\pm0.8}$ & 62.4$_{\pm1.5}$ & 44.2$_{\pm2.0}$ \\
        SST$_{AAA122}$  & 76.8$_{\pm4.2}$ & 53.6$_{\pm8.5}$ & 42.0$_{\pm0.5}$ & 74.9$_{\pm2.1}$ & 63.9$_{\pm1.6}$ & 29.0$_{\pm2.7}$ & 52.1$_{\pm1.3}$ & 12.4$_{\pm2.6}$ & \underline{81.3$_{\pm1.0}$} & \underline{63.5$_{\pm1.1}$} & 63.2$_{\pm0.9}$ & 46.7$_{\pm1.8}$ \\
        SARB$_{AAA122}$ & 63.4$_{\pm4.3}$ & 28.2$_{\pm7.4}$ & 41.7$_{\pm0.8}$ & 70.8$_{\pm3.3}$ & 65.8$_{\pm0.7}$ & 32.0$_{\pm1.8}$ & 57.4$_{\pm3.3}$ & \underline{18.9$_{\pm2.8}$} & 74.3$_{\pm2.6}$ & 51.3$_{\pm4.2}$ & 60.5$_{\pm1.0}$ & 40.3$_{\pm1.2}$ \\
        BoostLU$_{CVPR23}$  & 46.0$_{\pm1.2}$ & 1.2$_{\pm1.5}$ & 37.3$_{\pm1.2}$ & 58.1$_{\pm2.9}$ & 52.8$_{\pm1.9}$ & 13.7$_{\pm2.7}$ & 53.9$_{\pm1.8}$ & 10.3$_{\pm3.5}$ & 32.6$_{\pm3.8}$ & 7.2$_{\pm1.2}$ & 44.5$_{\pm0.8}$ & 29.8$_{\pm2.3}$ \\
        HST$_{IJCV24}$  & 74.6$_{\pm5.9}$ & 48.4$_{\pm11.5}$ & 42.9$_{\pm2.3}$ & 74.0$_{\pm5.9}$ & 64.7$_{\pm2.0}$ & 29.7$_{\pm3.8}$ & 50.9$_{\pm0.7}$ & 18.0$_{\pm0.6}$ & 78.8$_{\pm1.8}$ & 59.0$_{\pm3.2}$ & 62.3$_{\pm1.7}$ & 44.8$_{\pm3.8}$ \\
        CALDNR$_{TMM24}$ & 55.8$_{\pm4.4}$ & 14.9$_{\pm9.7}$ & 32.1$_{\pm6.3}$ & 42.1$_{\pm18.5}$ & 58.8$_{\pm1.1}$ & 22.3$_{\pm1.7}$ & 50.6$_{\pm3.2}$ & 3.3$_{\pm5.7}$ & 41.0$_{\pm3.7}$ & 12.5$_{\pm6.2}$ & 47.6$_{\pm3.1}$ & 19.0$_{\pm6.7}$ \\
        PSScreen V1$_{BMVC2025}$ & \underline{80.2$_{\pm1.4}$} & \underline{60.5$_{\pm2.5}$} & \textbf{46.7$_{\pm0.6}$} & \underline{84.5$_{\pm0.9}$} & 68.0$_{\pm1.2}$ & 36.9$_{\pm2.2}$ & \textbf{59.1$_{\pm2.6}$} & \textbf{19.6$_{\pm4.6}$} & 75.4$_{\pm0.8}$ & 53.1$_{\pm1.3}$ & \underline{65.9$_{\pm0.1}$} & \underline{50.9$_{\pm0.1}$} \\
        \textbf{PSScreen V2}  & \textbf{81.9$_{\pm1.7}$} & \textbf{64.2$_{\pm3.5}$} & \underline{46.4$_{\pm0.5}$} & \textbf{85.4$_{\pm0.8}$} & \underline{69.3$_{\pm0.7}$} & \underline{38.9$_{\pm1.2}$} & \underline{58.6$_{\pm2.1}$} & 18.1$_{\pm4.3}$ & 74.3$_{\pm2.7}$ & 51.3$_{\pm4.5}$ & \textbf{66.1$_{\pm0.8}$} & \textbf{51.6$_{\pm1.4}$} \\
         \botrule
    \end{tabular}}
    \label{tab:metrics comparison in Unseen}
\end{table*}

\textbf{Results on Unseen and ODIR.} We validate the generalization capability of screening models on two out-of-domain datasets and report performances on Unseen in Tab. \ref{tab:metrics comparison in Unseen} and on ODIR~\citep{ODIR_2021} in Tab.~\ref{tab:merged_f_qwk_odir}, respectively. We can observe that our \textbf{PSScreen V2}  significantly outperforms state-of-the-art methods on out-of-domain datasets. In detail, on Unseen, \textbf{PSScreen V2} achieves improvements of 0.2\% in $mF$ and 0.7\% in $mQWK$ over the second-best method, i.e., PSScreen V1 \citep{PSScreen_2025_BMVC}. More specifically, \textbf{PSScreen V2} achieves the best or second-best F-scores on four out of five classes in Unseen. The compared methods, except for PSScreen V1 \citep{PSScreen_2025_BMVC}, generally assume that the training and test data follow the same distributions, which limits their performances on out-of-domain data.

\begin{table*}[t!]
    \centering
    \Large
    \caption{F-score (\%) and QWK (\%) on ODIR \citep{ODIR_2021}. The best and second-best are highlighted in bold and with an underline.}
    \resizebox{\textwidth}{!}{
    \begin{tabular}{lcccccccccccccccc}
        \toprule
        Methods & \multicolumn{2}{c}{T1: Normal} & \multicolumn{2}{c}{T2: DR} & \multicolumn{2}{c}{T3: Glaucoma} & \multicolumn{2}{c}{T4: Cataract} & \multicolumn{2}{c}{T5: AMD} & \multicolumn{2}{c}{T6: HR} & \multicolumn{2}{c}{T7: PM} & \multicolumn{2}{c}{Average} \\
        \cmidrule(lr){2-3} \cmidrule(lr){4-5} \cmidrule(lr){6-7} \cmidrule(lr){8-9} \cmidrule(lr){10-11} \cmidrule(lr){12-13} \cmidrule(lr){14-15} \cmidrule(lr){16-17}
        & F-score & QWK & F-score & QWK & F-score & QWK & F-score & QWK & F-score & QWK & F-score & QWK & F-score & QWK & $mF$ & $mQWK$ \\
        \midrule
        \rowcolor{gray!10}
        Fully Supervised & 70.8$_{\pm0.2}$ & 41.6$_{\pm0.4}$ & 32.1$_{\pm1.7}$ & 53.6$_{\pm3.1}$ & 66.8$_{\pm0.9}$ & 33.8$_{\pm1.7}$ & 91.8$_{\pm0.8}$ & 83.5$_{\pm1.5}$ & 80.2$_{\pm1.4}$ & 60.5$_{\pm2.8}$ & 58.8$_{\pm1.1}$ & 19.2$_{\pm0.8}$ & 87.8$_{\pm1.2}$ & 75.6$_{\pm2.3}$ & 69.8$_{\pm0.2}$ & 52.6$_{\pm0.7}$ \\
        MultiNets & 59.3$_{\pm2.6}$ & 19.2$_{\pm4.4}$ & 28.7$_{\pm0.7}$ & 29.7$_{\pm1.6}$ & 60.8$_{\pm1.7}$ & 24.8$_{\pm2.2}$ & 66.2$_{\pm1.9}$ & 35.4$_{\pm3.2}$ & 59.2$_{\pm5.7}$ & 22.9$_{\pm8.3}$ & 47.7$_{\pm2.9}$ & \underline{3.9$_{\pm2.5}$} & 72.7$_{\pm1.4}$ & 46.6$_{\pm2.5}$ & 56.4$_{\pm1.4}$ & 26.1$_{\pm2.1}$ \\
        MultiHeads & 53.9$_{\pm1.1}$ & 17.9$_{\pm1.9}$ & 29.7$_{\pm1.3}$ & 34.0$_{\pm2.5}$ & 62.8$_{\pm2.4}$ & 26.9$_{\pm4.1}$ & 67.3$_{\pm0.5}$ & 36.5$_{\pm1.3}$ & 57.4$_{\pm2.4}$ & 18.8$_{\pm3.6}$ & 47.7$_{\pm1.1}$ & 3.2$_{\pm1.7}$ & 78.2$_{\pm1.9}$ & 56.8$_{\pm3.6}$ & 56.7$_{\pm0.6}$ & 27.7$_{\pm0.4}$ \\
        SST$_{AAAI2022}$ & 62.8$_{\pm3.2}$ & 28.9$_{\pm4.3}$ & 29.1$_{\pm1.6}$ & 32.3$_{\pm1.0}$ & 58.7$_{\pm1.0}$ & 21.7$_{\pm1.5}$ & 70.9$_{\pm3.9}$ & 43.3$_{\pm6.8}$ & 53.0$_{\pm1.1}$ & 14.0$_{\pm1.8}$ & 44.1$_{\pm1.5}$ & 0.3$_{\pm1.1}$ & 70.6$_{\pm4.2}$ & 42.8$_{\pm7.5}$ & 55.6$_{\pm1.7}$ & 26.2$_{\pm2.5}$ \\
        SARB$_{AAAI2022}$& 60.8$_{\pm2.2}$ & 24.3$_{\pm3.6}$ & 31.7$_{\pm2.3}$ & 31.9$_{\pm2.1}$ & 59.1$_{\pm1.9}$ & 21.0$_{\pm2.9}$ & 84.6$_{\pm2.2}$ & 69.3$_{\pm4.5}$ & 61.2$_{\pm4.2}$ & 24.5$_{\pm7.2}$ & 51.0$_{\pm1.3}$ & 2.6$_{\pm2.2}$ & 81.3$_{\pm0.9}$ &62.9$_{\pm1.8}$ & 61.4$_{\pm0.9}$ & 33.8$_{\pm1.6}$ \\
        BoostLU$_{CVPR2023}$& 36.1$_{\pm0.0}$ & 0.0$_{\pm0.0}$ & 29.6$_{\pm2.8}$ & \underline{39.6$_{\pm7.3}$} & 63.5$_{\pm3.4}$ & 24.7$_{\pm1.0}$ & 48.8$_{\pm0.0}$ & 0.0$_{\pm0.0}$ & 61.7$_{\pm5.5}$ & 24.5$_{\pm10.7}$ & 49.0$_{\pm0.0}$ & 0.0$_{\pm0.0}$ & 66.8$_{\pm3.1}$ & 34.0$_{\pm6.2}$ & 50.6$_{\pm1.0}$ & 17.5$_{\pm2.1}$ \\
        HST$_{IJCV2024}$& 60.9$_{\pm2.4}$ & 25.4$_{\pm2.9}$ & 29.0$_{\pm1.9}$ & 33.1$_{\pm0.4}$ & 60.4$_{\pm1.2}$ & 23.9$_{\pm1.5}$ & 76.0$_{\pm2.8}$ & 52.7$_{\pm5.2}$ & 50.4$_{\pm2.0}$ & 11.8$_{\pm1.9}$ & 43.2$_{\pm1.7}$ & 0.6$_{\pm0.4}$ & 71.9$_{\pm1.7}$ & 45.3$_{\pm3.2}$ & 56.0$_{\pm0.8}$ & 27.5$_{\pm0.8}$ \\
        CALDNR$_{TMM2024}$& 36.1$_{\pm0.0}$ & 0.0$_{\pm0.0}$ & \textbf{33.5$_{\pm3.1}$} & 38.7$_{\pm3.9}$ & 63.5$_{\pm3.4}$ & 27.3$_{\pm6.6}$ & 55.9$_{\pm10.6}$ & 13.5$_{\pm20.3}$ & 58.9$_{\pm3.7}$ & 18.7$_{\pm7.1}$ & 49.3$_{\pm0.0}$ & 0.0$_{\pm0.0}$ & 67.9$_{\pm1.8}$ & 36.1$_{\pm3.6}$ & 52.1$_{\pm0.6}$ & 19.1$_{\pm0.8}$ \\
        PSScreen V1$_{BMVC2025}$  & \underline{66.8$_{\pm1.4}$} & \underline{34.6$_{\pm2.1}$} & \underline{33.1$_{\pm1.9}$} & \textbf{44.8$_{\pm2.5}$} & \textbf{66.9$_{\pm2.3}$} & \textbf{34.0$_{\pm4.4}$} & \underline{85.9$_{\pm1.6}$} & \underline{71.9$_{\pm3.1}$} & \underline{63.0$_{\pm0.8}$} & \underline{27.6$_{\pm1.4}$} & \underline{51.1$_{\pm2.3}$} & 2.3$_{\pm4.6}$ & \textbf{81.6$_{\pm2.6}$} & \textbf{63.4$_{\pm5.2}$} & \underline{64.1$_{\pm1.0}$} & \underline{39.8$_{\pm1.3}$} \\
        \textbf{PSScreen V2} & \textbf{67.6$_{\pm0.8}$} & \textbf{35.5$_{\pm1.5}$} & 29.5$_{\pm1.2}$ & 36.8$_{\pm2.7}$ & \underline{66.2$_{\pm0.9}$} & \underline{33.1$_{\pm2.0}$} & \textbf{86.4$_{\pm2.1}$} & \textbf{72.8$_{\pm4.3}$} & \textbf{63.8$_{\pm1.0}$} & \textbf{29.1$_{\pm1.6}$} & \textbf{54.5$_{\pm1.1}$} & \textbf{9.2$_{\pm2.2}$} & \underline{81.5$_{\pm0.4}$} & \underline{63.2$_{\pm0.7}$} & \textbf{64.2$_{\pm0.4}$} & \textbf{39.9$_{\pm0.9}$} \\
        \botrule
    \end{tabular}
    }
    \label{tab:merged_f_qwk_odir}
\end{table*}

\begin{figure}[!t] 
    \centering 
    \includegraphics[width=0.4\columnwidth]{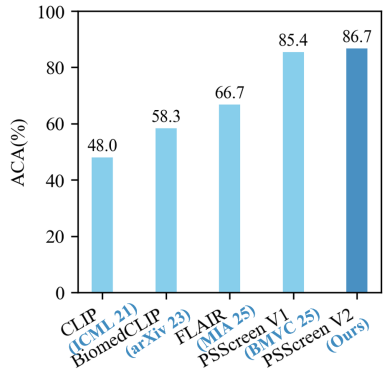} 
    \caption{Performance comparison of zero-shot inference with foundation models on the ODIR200$\times$3 dataset.} 
    \label{fig:zero-shot} 
\end{figure}

ODIR \citep{ODIR_2021} provides labels for all seven classes, thus we additionally report the performances by the fully supervised screening model as the upper bound in Tab. \ref{tab:merged_f_qwk_odir}. It can be observed that (1) \textbf{PSScreen V2} and PSScreen V1 \citep{PSScreen_2025_BMVC} rank first and second, and they significantly outperform the compared partially supervised methods; (2) Compared to the fully supervised model, significant potential for improvement remains.

\subsection{Partially Supervised Learning or Zero-shot Inference with Foundation Model?}
We further compare our \textbf{PSScreen V2} and PSScreen V1 \citep{PSScreen_2025_BMVC} with three  foundation models, i.e., CLIP~\citep{CLIP_2021_ICML}, BiomedCLIP~\citep{BiomedCLIP_2023_arXiv}, and FLAIR~\citep{FLAIR_2025_MIA} on ODIR200$\times$3 under the setting of zero-shot inference. Following FLAIR, we adopt ACA~\citep{ACA_2019_ICIP} as the evaluation metric and report ACAs in Fig.~\ref{fig:zero-shot}. As shown, foundation models perform suboptimally in zero-shot retinal disease screening and fail to generalize to unseen diseases due to the complex pathological structures of retinal diseases. With the supervision from partially labelled samples during training, our \textbf{PSScreen V2} and V1 \citep{PSScreen_2025_BMVC} significantly outperform the best foundation model FLAIR by 20\% and 18.7\% respectively. This demonstrates the superiority of the partially supervised learning paradigm, which fully leverages both images and annotations compared to the self-supervised learning paradigm in disease screening.

\begin{table*}[t!]
\centering
\caption{$mF$ (\%) and $mQWK$ (\%) by methods using DINOv2 backbone.}
\resizebox{0.8\textwidth}{!}{
\begin{tabular}{lcccccc}
\toprule
Methods & \multicolumn{2}{c}{Meta} & \multicolumn{2}{c}{Unseen} & \multicolumn{2}{c}{ODIR}   \\
\cmidrule(lr){2-3}\cmidrule(lr){4-5}\cmidrule(lr){6-7}
 & $mF$ & $mQWK$ & $mF$ & $mQWK$ & $mF$ & $mQWK$ \\
\midrule
MultiNets & 81.8$_{\pm0.4}$ & 71.1$_{\pm0.5}$ & 61.8$_{\pm0.8}$ & 42.3$_{\pm1.3}$ & 52.4$_{\pm0.6}$ & 22.4$_{\pm1.0}$ \\
MultiHeads & 82.8$_{\pm0.4}$ & 74.3$_{\pm1.1}$ & 63.0$_{\pm0.4}$ & 45.8$_{\pm0.9}$ & 54.0$_{\pm1.0}$ & 23.8$_{\pm1.0}$  \\
SST$_{AAA122}$ & 78.5$_{\pm0.8}$ & 67.1$_{\pm1.8}$ & 60.8$_{\pm0.5}$ & 41.7$_{\pm1.2}$ & 50.4$_{\pm0.8}$ & 19.6$_{\pm1.7}$  \\
SARB$_{AAA122}$ & 77.3$_{\pm0.7}$ & 65.1$_{\pm1.6}$ & 59.7$_{\pm0.8}$ & 39.3$_{\pm1.8}$ & 52.8$_{\pm1.6}$ & 21.6$_{\pm1.6}$  \\
HST$_{IJCV24}$ & 77.4$_{\pm0.1}$ & 65.5$_{\pm0.3}$ & 61.3$_{\pm1.1}$ & 41.9$_{\pm0.7}$ & 51.3$_{\pm1.0}$ & 21.0$_{\pm1.2}$  \\
BoostLU$_{CVPR23}$& 79.7$_{\pm1.3}$ & 68.9$_{\pm1.6}$ & 48.7$_{\pm1.4}$ & 21.5$_{\pm1.1}$ & 50.9$_{\pm1.6}$ & 18.3$_{\pm1.2}$  \\
CALDNR$_{TMM24}$ & 80.9$_{\pm0.5}$ & 71.2$_{\pm1.1}$ & 51.2$_{\pm0.8}$ & 22.5$_{\pm2.5}$ & 52.5$_{\pm1.0}$ & 21.0$_{\pm2.2}$ \\
PSScreen V1$_{BMVC25}$ & \underline{83.1$_{\pm0.2}$} & \underline{75.7$_{\pm0.7}$}  & \underline{63.3$_{\pm1.5}$} & \underline{45.9$_{\pm2.2}$} & \underline{58.0$_{\pm0.5}$} & \underline{27.9$_{\pm1.4}$}  \\
\textbf{PSScreen V2} & \textbf{83.8$_{\pm0.2}$} & \textbf{76.0$_{\pm0.7}$}  & \textbf{65.3$_{\pm1.0}$} & \textbf{49.3$_{\pm1.6}$} & \textbf{59.8$_{\pm1.1}$} & \textbf{31.2$_{\pm1.1}$}  \\
\botrule
\end{tabular}
}
\label{tab:screening_dinov2}
\end{table*}%

\subsection{PSScreen V2 with DINOv2 as the Backbone} 
To further explore the potential of \textbf{PSScreen V2} in enhancing the performance of vision foundation model (VFM) for multiple retinal disease screening, we replace the backbone ResNet-101 \citep{ResNet_2016_CVPR} with DINOv2~\citep{DINOv2_2024_TMLR} for comparison. Specifically, we freeze all the parameters of DINOv2~\citep{DINOv2_2024_TMLR} and employ SoMA~\citep{SoMA_2025_CVPR} to adapt DINOv2 \citep{DINOv2_2024_TMLR} for retinal disease screening. SoMA~\citep{SoMA_2025_CVPR} is applied to all linear layers within the self-attention and MLP components of the last four blocks of DINOv2~\citep{DINOv2_2024_TMLR}, with a rank of 8. The initial learning rate is set to 1×10$^{\text{-3}}$ and the weight decay to 1×10$^{\text{-5}}$, while all other settings remain same with those used for ResNet-101. Similar changes are applied to the compared state-of-the-art methods and performances are reported in Tab.~\ref{tab:screening_dinov2}. Experimental results show that our \textbf{PSScreen V2} outperforms state-of-the-art methods significantly across three datasets. Compared with PSScreen V1 \citep{PSScreen_2025_BMVC}, \textbf{PSScreen V2} improves $mQWK$ by 0.3\%, 3.4\%, and 3.3\% across three datasets, with an average gain of 2.3\%.

\subsection{Ablation Study}
\textbf{Loss Weight Selection.} To determine the loss weight $\lambda_1$ of $\mathcal{L}^{unknown}_{\mathcal{S}_2\text{-}CE}$ in Eq. (\ref{Eq:Loss-S2-CE}), $\lambda_2$ of $\mathcal{L}_{MMD}$ and $\lambda_3$ of $\mathcal{L}_{KL}^{known}$ in Eq. (\ref{eq:overall_loss_S2}), we vary them and report $mQWK$ on the Meta validation set in Tab.~\ref{tab:lambda_search}, which shows that setting $\lambda_1=0.6$, $\lambda_2=0.05$, and $\lambda_3=1.0$ yields the best overall screening performance.

\begin{table*}[t!]
\centering
\caption{$mQWK$ (\%) on Val of Meta under different $\lambda_1$, $\lambda_2$, $\lambda_3$.}
\resizebox{0.25\textwidth}{!}{
\begin{tabular}{cccc}
 \toprule
 $\lambda_1$ & $\lambda_2$ & $\lambda_3$ & Meta \\
 \midrule
 0.6  &0.1   & 1.0    &  75.0$_{\pm0.6}$\\
 \textbf{0.6} & \textbf{0.05}  & \textbf{1.0}    & \textbf{75.4$_{\pm0.8}$}\\
 0.6 &0.025 & 1.0    &  74.9$_{\pm0.2}$\\
 \midrule
 0.6 &0.05  & 0.5    &  75.1$_{\pm1.1}$\\
 \textbf{0.6}  & \textbf{0.05}  & \textbf{1.0}    &  \textbf{75.4$_{\pm0.8}$}\\
 0.6 &0.05  & 2.0    &  74.8$_{\pm0.8}$\\
 \midrule
 0.4  & 0.05  & 1.0  & 74.7$_{\pm0.2}$\\
 \textbf{0.6} & \textbf{0.05}  & \textbf{1.0}    &  \textbf{75.4$_{\pm0.8}$}\\
 0.8  &0.05  & 1.0   & 75.0$_{\pm0.1}$ \\
 \botrule
\end{tabular}
}
\label{tab:lambda_search}
\end{table*}

\textbf{How Does Each Branch in \textbf{PSScreen V2} Contribute?} PSScreen V2 consists of three branches: $\mathcal{T}$, $\mathcal{S}_1$, and $\mathcal{S}_2$. To investigate their contributions, we conduct ablation experiments and report the performance in Table.~\ref{tab:ablation_branch}. Retaining only $\mathcal{T}$, \textbf{PSScreen V2} degrades into MultiHeads equipped with the text-guided semantic decoupling, which achieves inferior performance to PSScreen V2 obviously. $\mathcal{S}_1$ (LF-Dropout) applies a conservative augmentation that largely preserves the original domain information. This may explain its notable improvement over the $\mathcal{T}$-only baseline on the Meta, as well as good generalization to unseen domains. Differently, $\mathcal{S}_2$ (LF-Uncert) explores a broader augmentation space by modelling low-frequency uncertainty, which significantly improves out-of-domain detection but slightly compromises in-domain performance in terms of $mF$. When combined, \textbf{PSScreen V2} achieves the best in both in-domain and out-of-domain datasets.

\textbf{How Does Each Loss Term in PSScreen V2 Contribute?} There are four loss terms, i.e., $\mathcal{L}^{unknown}_{\mathcal{S}_1\text{-}CE}$ in Eq.~(\ref{eq:psl_AD}), $\mathcal{L}^{unknown}_{\mathcal{S}_2\text{-}CE}$ in Eq.~(\ref{eq:L_S2_CE_unknown}), $L_{MMD}$ in Eq.~(\ref{eq:MMD_loss}), and $\mathcal{L}_{KL}^{known}$ in Eq.~(\ref{eq:sd_label_loss}), except for the classification losses for labelled classes. To investigate how the loss items contribute, we ablate each individual loss term. As shown in Tab.~\ref{tab:ablation_psscreenv2}, without $\mathcal{L}^{unknown}_{\mathcal{S}_1\text{-}CE}$ or $\mathcal{L}^{unknown}_{\mathcal{S}_2\text{-}CE}$, the performances on three datasets consistently and significantly decrease, indicating that supervision from pseudo labels of unknown classes contributes substantially. Without $\mathcal{L}_{MMD}$ or $\mathcal{L}_{KL}^{known}$, performances degrade significantly. The performance degradation without $\mathcal{L}_{MMD}$ is more severe than that without $\mathcal{L}_{KL}^{known}$. The likely reason is that $\mathcal{L}_{MMD}$ facilitates knowledge transfer for both known and unknown classes via feature consistency, whereas $\mathcal{L}_{KL}^{known}$ restricts transfer to known classes via prediction consistency. Therefore, $\mathcal{L}_{MMD}$ contributes more than $\mathcal{L}_{KL}^{known}$. 

\begin{table*}[!t]
    \centering
    \large
    \caption{The ablation study on each branch of PSScreen V2.}
    \resizebox{0.8\textwidth}{!}{
    \begin{tabular}{lcccccc}
        \toprule
        Methods & \multicolumn{2}{c}{Meta} & \multicolumn{2}{c}{Unseen} & \multicolumn{2}{c}{ODIR} \\
        \cmidrule(lr){2-3} \cmidrule(lr){4-5} \cmidrule(lr){6-7}
         & $mF$ & $mQWK$ & $mF$ & $mQWK$ & $mF$ & $mQWK$  \\
        \midrule
        $\mathcal{T}$         & 83.6$_{\pm0.7}$ & 75.0$_{\pm1.3}$ & 62.4$_{\pm1.5}$ & 44.2$_{\pm2.0}$ & 56.7$_{\pm0.6}$ & 27.7$_{\pm0.4}$ \\
        $\mathcal{T} + \mathcal{S}_1$           & 84.2$_{\pm0.4}$ & 77.1$_{\pm0.7}$ & 63.3$_{\pm0.6}$ & 46.4$_{\pm1.1}$ & 60.8$_{\pm0.3}$ & 33.3$_{\pm0.5}$ \\
        $\mathcal{T} + \mathcal{S}_2$    & 83.2$_{\pm0.2}$ & 75.7$_{\pm0.4}$ & 65.2$_{\pm0.4}$ & 50.1$_{\pm0.6}$ & 62.3$_{\pm0.4}$ & 37.0$_{\pm1.0}$  \\
        $\mathcal{T} + \mathcal{S}_1 + \mathcal{S}_2$     & \textbf{84.5$_{\pm0.5}$} & \textbf{78.0$_{\pm1.0}$} & \textbf{66.1$_{\pm0.8}$} & \textbf{51.6$_{\pm1.4}$} & \textbf{64.2$_{\pm0.4}$} & \textbf{39.9$_{\pm0.9}$} \\
        \botrule
    \end{tabular}
    }
    \label{tab:ablation_branch}
\end{table*}

\begin{table*}[!t]
    \centering
    \large
    \caption{The ablation study on the key loss terms of PSScreen V2.}
    \resizebox{0.8\textwidth}{!}{
    \begin{tabular}{lcccccc}
        \toprule
        Methods & \multicolumn{2}{c}{Meta} & \multicolumn{2}{c}{Unseen} & \multicolumn{2}{c}{ODIR} \\
        \cmidrule(lr){2-3} \cmidrule(lr){4-5} \cmidrule(lr){6-7}
         & $mF$ & $mQWK$ & $mF$ & $mQWK$ & $mF$ & $mQWK$ \\
        \midrule
        \textbf{PSScreen V2}            & \textbf{84.5$_{\pm0.5}$} & \textbf{78.0$_{\pm1.0}$} & \textbf{66.1$_{\pm0.8}$} & \textbf{51.6$_{\pm1.4}$} & \textbf{64.2$_{\pm0.4}$} & \textbf{39.9$_{\pm0.9}$} \\
        w/o $\mathcal{L}^{unknown}_{\mathcal{S}_1\text{-}CE}$ & 83.2$_{\pm0.2}$ & 75.7$_{\pm0.4}$ & 65.2$_{\pm0.4}$ & 50.1$_{\pm0.6}$ & 62.3$_{\pm0.4}$ & 37.0$_{\pm1.0}$ \\
        w/o $\mathcal{L}^{unknown}_{\mathcal{S}_2\text{-}CE}$ & 83.7$_{\pm0.9}$ & 76.3$_{\pm1.8}$ & 64.7$_{\pm0.7}$ & 49.0$_{\pm0.9}$ & 61.3$_{\pm0.7}$ & 34.9$_{\pm1.0}$ \\
        w/o $\mathcal{L}_{MMD}$         & 83.8$_{\pm0.2}$ & 76.4$_{\pm0.5}$ & 65.2$_{\pm0.3}$ & 49.8$_{\pm0.6}$ & 62.4$_{\pm0.7}$ & 36.5$_{\pm1.1}$ \\
        w/o $\mathcal{L}_{KL}^{known}$  & 84.2$_{\pm1.0}$ & 77.2$_{\pm1.8}$ & 64.8$_{\pm1.0}$ & 49.5$_{\pm1.5}$ & 63.3$_{\pm0.8}$ & 38.4$_{\pm1.2}$ \\
        \botrule
    \end{tabular}
    }
    \label{tab:ablation_psscreenv2}
\end{table*}

\begin{figure*}[!t]
    \centering
    \includegraphics[width=0.4\linewidth]{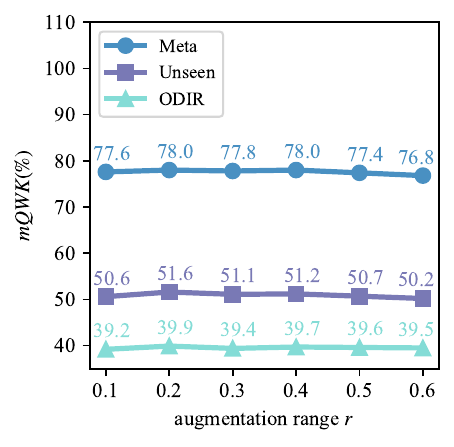}
    \caption{$mQWK$ under varying augmentation range $r$.}
    \label{fig:side_length}
\end{figure*}

\textbf{Ablation Study on the Augmentation Range $r$.} $r$ determines the range of low-frequency components for LF-Dropout and LF-Uncert augmentation. The larger the $r$, the wider the range. We vary $r$ from 0.1 to 0.6 with a step size of 0.1. The optimal value of $r$ is selected based on the Meta validation set, where $r=0.2$ yields the best performance; this setting is adopted as the default in all experiments unless otherwise specified. To illustrate the impact of varying $r$ on each dataset, we also  report the results on the test sets of three datasets in Fig.~\ref{fig:side_length}. It shows that increasing $r$ beyond 0.2 generally degrades performance, suggesting that overly disrupting high-frequency components may remove disease-related information.

\begin{figure*}[!t]
    \centering
    \includegraphics[width=0.4\linewidth]{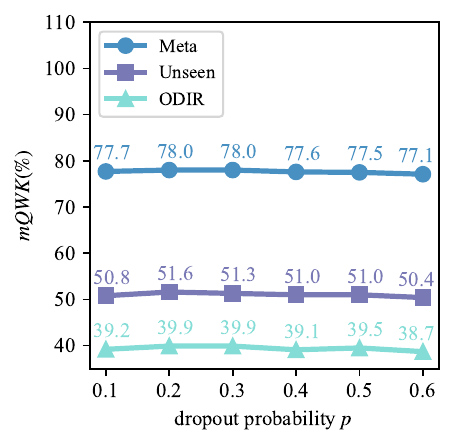}
    \caption{$mQWK$ under varying dropout probability $p$.}
    \label{fig:drop_prob}
\end{figure*}

\textbf{Ablation Study on LF-Dropout Probability $p$.}
\textbf{PSScreen V2} achieves the best performance on the Meta validation set when the $p$ is 0.2, which is adopted as the default setting in all experiments if not specified. To illustrate the impact of varying $p$ on each dataset, we also  report the results on the test sets of three datasets in Fig.~\ref{fig:drop_prob}. It shows that an excessively large $p$ results in the removal of too much low-frequency signals, which significantly degrades model performance.

\begin{table*}[!t]
\centering
\centering
\caption{$mQWK$ (\%) by methods using different backbones.}
\resizebox{0.8\textwidth}{!}{
\begin{tabular}{lcccc}
\toprule
Methods & Meta & Unseen & ODIR & Average \\
\midrule
ResNet-101~\citep{ResNet_2016_CVPR} + MultiHeads                     & 75.0$_{\pm1.3}$ & 44.2$_{\pm2.0}$ & 27.7$_{\pm0.4}$ & 49.0 \\
ResNet-101~\citep{ResNet_2016_CVPR} + PSScreen V1 & 76.8$_{\pm0.8}$ & 50.9$_{\pm0.1}$ & 39.8$_{\pm1.3}$ & 55.8 \\
\textbf{ResNet-101~\citep{ResNet_2016_CVPR}+ PSScreen V2}           & \textbf{78.0$_{\pm1.0}$} & \textbf{51.6$_{\pm1.4}$} & \textbf{39.9$_{\pm0.9}$} & \textbf{56.5} \\
\midrule
ConvNeXt V2-T~\citep{ConvNeXt_V2_2023_CVPR} + MultiHeads             & 76.3$_{\pm0.7}$ & 48.6$_{\pm1.2}$ & 28.9$_{\pm1.3}$ & 51.3 \\
ConvNeXt V2-T~\citep{ConvNeXt_V2_2023_CVPR}+ PSScreen V1 & 77.0$_{\pm0.2}$ & \textbf{49.1$_{\pm0.5}$} & 38.3$_{\pm1.0}$ & 54.8 \\
\textbf{ConvNeXt V2-T~\citep{ConvNeXt_V2_2023_CVPR} + PSScreen V2}  & \textbf{78.6$_{\pm0.8}$} & 49.1$_{\pm0.6}$ & \textbf{38.9$_{\pm0.8}$} & \textbf{55.5} \\
\midrule
VMamba-T~\citep{VMamba_2024_NeurIPS} + MultiHeads              & 76.3$_{\pm1.0}$ & 47.2$_{\pm0.5}$ & 27.0$_{\pm0.4}$ & 50.2 \\
VMamba-T~\citep{VMamba_2024_NeurIPS} + PSScreen V1 & 78.0$_{\pm0.7}$ & \textbf{49.8$_{\pm1.7}$} & 36.9$_{\pm1.8}$ & 54.9 \\
\textbf{VMamba-T~\citep{VMamba_2024_NeurIPS} + PSScreen V2}      & \textbf{79.0$_{\pm0.4}$} & 49.2$_{\pm2.5}$ & \textbf{39.0$_{\pm1.0}$} & \textbf{55.7} \\
\midrule
Swin-T~\citep{Swin_2021_ICCV} + MultiHeads              & 75.8$_{\pm0.8}$ & 49.5$_{\pm0.7}$ & 25.4$_{\pm0.3}$ & 50.2 \\
Swin-T~\citep{Swin_2021_ICCV}+ PSScreen V1 & \textbf{79.3$_{\pm0.7}$} & \textbf{50.8$_{\pm0.6}$} & \textbf{35.5$_{\pm0.9}$} & \textbf{55.2} \\
\textbf{Swin-T~\citep{Swin_2021_ICCV}+ PSScreen V2}       & \textbf{79.3$_{\pm0.1}$} & 47.0$_{\pm1.3}$ & \textbf{35.5$_{\pm0.9}$} & 53.9 \\
\botrule
\end{tabular}
}
\label{tab:diff_backbone}
\end{table*}

\begin{table*}[]
\centering
\caption{$mQWK$ (\%) by models with DINOv2 \citep{DINOv2_2024_TMLR} using different PEFT methods.}
\resizebox{0.8\textwidth}{!}{
\begin{tabular}{lcccc}
\toprule
Methods & Meta & Unseen & ODIR & Average \\
\midrule
DoRA \citep{DoRA_2024_ICML} + MultiHeads                 & \textbf{74.5$_{\pm0.6}$} & 47.1$_{\pm1.2}$ & 22.4$_{\pm1.0}$ & 48.0 \\
DoRA  \citep{DoRA_2024_ICML} + PSScreen V1& 73.5$_{\pm1.5}$ & 46.8$_{\pm1.7}$ & 31.9$_{\pm0.5}$ & 50.7 \\
\textbf{DoRA  \citep{DoRA_2024_ICML} + PSScreen V2}        & 74.0$_{\pm2.4}$ & \textbf{48.9$_{\pm0.6}$} & \textbf{32.6$_{\pm2.5}$} & \textbf{51.8} \\
\midrule
FLoRA \citep{FLoRA_2025_ICLR} + MultiHeads              & 73.8$_{\pm0.7}$ & 43.6$_{\pm1.0}$ & 21.6$_{\pm1.6}$ & 46.3 \\
FLoRA \citep{FLoRA_2025_ICLR} + PSScreen V1 & 75.5$_{\pm1.2}$ & 45.1$_{\pm1.0}$ & 32.5$_{\pm0.4}$ & 51.0 \\
\textbf{FLoRA \citep{FLoRA_2025_ICLR} + PSScreen V2}      & \textbf{76.1$_{\pm1.3}$} & \textbf{47.1$_{\pm0.3}$} & \textbf{32.9$_{\pm0.9}$} & \textbf{52.0} \\
\midrule
LoRA-Dash \citep{LoRA_Dash_2025_ICLR} + MultiHeads         & 74.7$_{\pm1.2}$ & 43.6$_{\pm1.3}$ & 21.0$_{\pm2.2}$ & 46.4 \\
LoRA-Dash \citep{LoRA_Dash_2025_ICLR} + PSScreen V1 & 73.9$_{\pm0.8}$ & \textbf{50.0$_{\pm1.0}$} & 25.6$_{\pm0.3}$ & 49.8 \\
\textbf{LoRA-Dash \citep{LoRA_Dash_2025_ICLR} + PSScreen V2} & \textbf{75.9$_{\pm2.3}$} & 46.8$_{\pm0.4}$ & \textbf{33.2$_{\pm1.8}$} & \textbf{52.0} \\
\midrule
SoMA  \citep{SoMA_2025_CVPR} + MultiHeads                 & 74.3$_{\pm1.1}$ & 45.8$_{\pm0.9}$ & 21.0$_{\pm2.2}$ & 47.0 \\
SoMA  \citep{SoMA_2025_CVPR} + PSScreen V1 & 75.7$_{\pm0.7}$ & 45.9$_{\pm2.2}$ & 27.9$_{\pm1.4}$ & 49.8 \\
\textbf{SoMA \citep{SoMA_2025_CVPR} + PSScreen V2}        & \textbf{76.0$_{\pm0.7}$} & \textbf{49.3$_{\pm1.6}$} & \textbf{31.2$_{\pm1.1}$} & \textbf{52.2} \\
\botrule
\end{tabular}
}
\label{tab:diff_peft}
\end{table*}
\textbf{Compatibility of PSScreen V2.} We investigate the compatibility of PSScreen V2 using (1) various pretrained backbones including ResNet-101~\citep{ResNet_2016_CVPR}, ConvNeXt V2-T~\citep{ConvNeXt_V2_2023_CVPR}, Swin-T~\citep{Swin_2021_ICCV}, and VMamba-T~\citep{VMamba_2024_NeurIPS}, and (2) DINOv2~\citep{DINOv2_2024_TMLR} fine-tuned using various parameter-efficient fine-tuning (PEFT) strategies including DoRA~\citep{DoRA_2024_ICML}, FLoRA~\citep{FLoRA_2025_ICLR}, LoRA-Dash~\citep{LoRA_Dash_2025_ICLR}, and SoMA~\citep{SoMA_2025_CVPR}. We compare our \textbf{PSScreen V2} with the baseline MultiHeads and PSScreen V1 \citep{PSScreen_2025_BMVC}. Performances of methods using various backbones are reported in Tab.~\ref{tab:diff_backbone}, which clearly demonstrate that \textbf{PSScreen V2} is compatible with various backbones and improves $mQWK$ across all three datasets in most cases compared to baseline and PSScreen V1 \citep{PSScreen_2025_BMVC}. Performances of methods with DINOv2~\cite{DINOv2_2024_TMLR} using various PEFT strategies are reported in Tab. \ref{tab:diff_peft}. Compared with the baseline MultiHeads and PSScreen V1 \citep{PSScreen_2025_BMVC}, \textbf{PSScreen V2} achieves better in most cases on the three datasets.

\textbf{Superiority over Existing Feature Augmentations.} To validate the superiority of our LF-Uncert, we replace LF-Uncert in \textbf{PSScreen V2} with six state-of-the-art feature augmentation strategies including StyleAdv~\citep{StyleAdv_2023_CVPR}, CSU~\citep{CSU_2024_WACV}, CAIC~\citep{FSC_2025_TIP}, AdvStyle~\citep{AdvStyle_2023_arXiv}, ALOFT~\citep{ALOFT_2023_CVPR}, and DSU~\citep{DSU_2022_ICLR}. Among these, ALOFT \citep{ALOFT_2023_CVPR} augments features in the frequency domain while the others operate in spatial domain. We report the performances in Tab. \ref{tab:feature_perturb_comparison}. As shown, compared to ALOFT \citep{ALOFT_2023_CVPR}, our LF-Uncert achieves improvements by 1.2\% on Meta, 2.0\% on Unseen and 2.1\% on ODIR \citep{ODIR_2021}. The possible reason is that our LF-Uncert learns distribution parameters for channel statistics adversarially which enables a broader augmentation space than ALOFT~\citep{ALOFT_2023_CVPR}, which estimates these parameters from batch samples. Compared to spatial-domain feature augmentation strategies, LF-Uncert surpasses the second-best methods: CAIC~\citep{FSC_2025_TIP} (+1.0\% on Meta), AdvStyle~\citep{AdvStyle_2023_arXiv} (+0.7\% on Unseen), and DSU~\citep{DSU_2022_ICLR} (+1.0\% on ODIR). 

\begin{table*}[t!]
\centering
\caption{$mQWK$ (\%) under different feature augmentation methods.}
\resizebox{0.7\textwidth}{!}{
\begin{tabular}{lccc}
\toprule
Methods & Meta & Unseen & ODIR \\
\midrule
StyleAdv$_{CVPR23}$~\citep{StyleAdv_2023_CVPR} & 76.1$_{\pm1.5}$ & 50.8$_{\pm0.6}$ & 33.2$_{\pm0.9}$ \\
CSU$_{WACV24}$~\citep{CSU_2024_WACV} & 74.7$_{\pm0.7}$ & 50.1$_{\pm1.1}$ & 36.8$_{\pm0.8}$ \\
CAIC$_{TIP25}$~\citep{FSC_2025_TIP} & 77.0$_{\pm0.8}$ & 50.4$_{\pm1.4}$ & 33.0$_{\pm1.1}$ \\
AdvStyle$_{arXiv23}$~\citep{AdvStyle_2023_arXiv} & 74.6$_{\pm1.5}$ & 50.9$_{\pm1.1}$ & 36.7$_{\pm1.4}$ \\
DSU$_{ICLR22}$~\citep{DSU_2022_ICLR} & 76.0$_{\pm0.9}$ & 49.5$_{\pm1.8}$ & 38.9$_{\pm1.8}$ \\
ALOFT$_{CVPR23}$~\citep{ALOFT_2023_CVPR} & 76.8$_{\pm0.5}$ & 49.6$_{\pm1.4}$ & 37.8$_{\pm1.9}$ \\
\textbf{LF-Uncert (Ours)} & \textbf{78.0$_{\pm1.0}$} & \textbf{51.6$_{\pm1.4}$} & \textbf{39.9$_{\pm0.9}$} \\
\botrule
\end{tabular}
}
\label{tab:feature_perturb_comparison}
\end{table*}

\begin{table*}[t!]

\centering
\caption{Complexity comparison of PSScreen V2 with other methods.}
\resizebox{0.5\textwidth}{!}{
\begin{tabular}{lccc}
\toprule
Methods & GFLOPs & \#param (M) & FPS \\
\midrule
MultiNets & 3928 & 1019.8 & 57.3 \\
MultiHeads & 655 & 170.0 & 381.1 \\
SARB& 666 & 217.8 & 323.5 \\
PSScreen V1& 661 & 177.8 & 372.0 \\
\textbf{PSScreen V2} & \textbf{661} & \textbf{177.9} & \textbf{371.6} \\
\botrule
\end{tabular}
}
\label{tab:model_cost_comparison}
\end{table*}

\subsection{Computational Cost}
To assess the computational efficiency, we compare \textbf{PSScreen V2} with the baseline method (MultiNets and MultiHeads) and the two state-of-the-art methods, i.e., SARB \citep{SARB_2022_AAAI} and PSScreen V1 \citep{PSScreen_2025_BMVC}, and report the GFLOPs, the total parameters $\#param$ and FPS (image per second) in Tab.~\ref{tab:model_cost_comparison}. We can observe that \textbf{PSScreen V2} and PSScreen V1 exhibit comparable performance in terms of GFLOPs, $\#param$, and FPS, and both consistently outperform the state-of-the-art method SARB~\citep{SARB_2022_AAAI}, being only slightly inferior to the naive baseline MultiHeads. This demonstrates that our \textbf{PSScreen V2} achieves superior performance without introducing obvious additional computational cost.

\begin{figure*}[t!]
    \centering
        \centering
        \includegraphics[width=\linewidth]{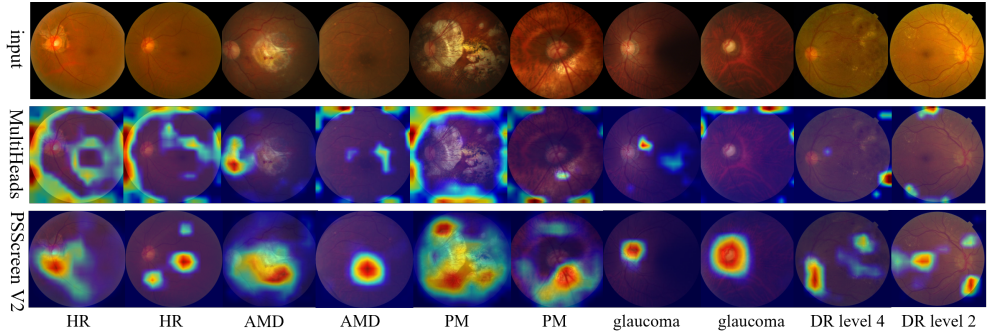}
        \caption{Visualization of heatmaps generated by different models for five retinal diseases: hypertensive retinopathy (HR), age-related macular degeneration (AMD), pathologic myopia (PM), glaucoma, and diabetic retinopathy (DR).}
        \label{fig:visualization}
\end{figure*}

\begin{figure*}[t!]
        \centering
        \includegraphics[width=0.4\linewidth]{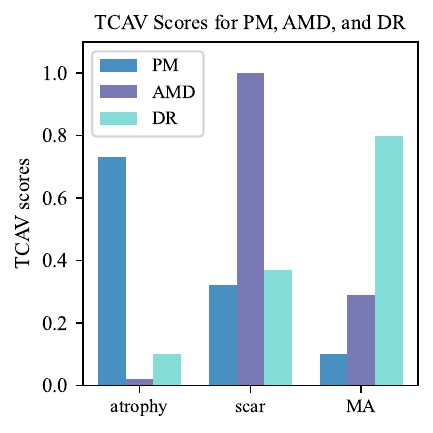}
        \caption{TCAV analysis \citep{TCAV_2018_ICML} for PM, AMD, and DR screening.}
        \label{fig:TCAV}
\end{figure*}

\subsection{Visualization and Interpretability Analysis}
To gain insights into the effectiveness of \textbf{PSScreen V2}, we use Grad-CAM~\citep{Grad-CAM_2017_ICCV} to visualise the informative regions in feature maps of the backbone's last convolutional layer for both MultiHeads and \textbf{PSScreen V2}. Fig.~\ref{fig:visualization} shows the Grad-CAM~\citep{Grad-CAM_2017_ICCV} heatmaps for 10 examples with five diseases: hypertensive retinopathy (HR), age-related macular degeneration (AMD), pathologic myopia (PM), glaucoma, and diabetic retinopathy (DR). As shown, MultiHeads tends to focus on background regions irrelevant to the diseases, while \textbf{PSScreen V2} usually attends to lesion/abnormal areas associated with diseases. Specifically, in examples of HR, \textbf{PSScreen V2} localizes the disease manifestation regions where appear arterio-venous crossing changes. In examples of AMD, \textbf{PSScreen V2} can accurately attend to regions where appear drusen, scar, and hemorrhages. In the PM cases, \textbf{PSScreen V2} focuses on patchy retinal atrophy typically associated with PM. In glaucoma cases, \textbf{PSScreen V2} correctly emphasizes the optic disc and optic cup regions. In DR cases, \textbf{PSScreen V2} focuses on lesions such as hemorrhages, hard exudates, and microaneurysms.

We employ TCAV analysis~\citep{TCAV_2018_ICML} to quantitively assess whether \textbf{PSScreen V2} relies on clinically meaningful pathological concepts for screening. Specifically, TCAV encodes each concept as a Concept Activation Vector (CAV) in the hidden space and quantifies its influence on the model’s prediction for a target class. In our experiments, we use “patchy retinal atrophy,” “scar,” and “microaneurysms (MA)” as concepts for PM, AMD, and DR, respectively. The corresponding datasets are PALM~\citep{PALM_2024} for PM, ADAM~\citep{ADAM_2022_TMI} for AMD, and DDR~\citep{DDR_2019} for DR. For each case, CAVs are constructed from the validation set, and TCAV scores are then computed on the test set. As shown in Fig.~\ref{fig:TCAV}, PSScreen V2 shows strong alignment between pathological concepts and corresponding diseases, confirming that it leverages clinically meaningful features and provides concept-level interpretability.

\section{Experiments on Chest X-ray Image Classification}
\textbf{Datasets and Implementation Details.} We conduct experiments on CheXpert~\citep{Chexpert_2019_AAAI}, SIIM-ACR Pneumothorax2019~\footnote{\href{https://www.kaggle.com/c/siim-acr-pneumothorax-segmentation}{SIIM-ACR Pneumothorax2019}\label{SIIM-ACR}} and ChestX-ray14~\citep{chestXray14_2017_CVPR}. \textbf{CheXpert}~\citep{Chexpert_2019_AAAI} contains 223,414 partially labelled images for training, 234 and 667 fully labelled image for validation and test, covering 14 classes. \textbf{SIIM-ACR Pneumothorax2019}~\footref{SIIM-ACR} includes 10,675 training and 1,372 test images with binary labels for pneumothorax detection. \textbf{ChestX-ray14}~\citep{chestXray14_2017_CVPR} contains 86,524 training and 25,596 test images, each fully annotated for 14 classes. In our protocol, we focus on eight overlapping classes between CheXpert~\citep{Chexpert_2019_AAAI} and ChestX-ray14~\citep{chestXray14_2017_CVPR}: Normal (Norm), Cardiomegaly (Card), Edema (Edem), Consolidation (Cons), Pneumonia (Pneu1), Atelectasis (Atel), Pneumothorax (Pneu2), and Effusion (Effu). CheXpert~\citep{Chexpert_2019_AAAI} is used for model training and in-domain testing, while SIIM-ACR Pneumothorax2019~\footref{SIIM-ACR} and ChestX-ray14~\citep{chestXray14_2017_CVPR} are used for out-of-domain testing.

We follow~\citep{MPC_2024_TMI} and resize chest X-ray images to 224$\times$224. For data augmentation, we use horizontal and vertical flips with a probability of 0.5, rotation within $\pm 30^\circ$, and Gaussian blur. We use pretrained ResNet-101 as the backbone and set the batch size to 64 during training and maintain the same training configurations as used for retinal disease screening. Macro-averaging F-score is used as the evaluation metric. We compare \textbf{PSScreen V2} with the baseline MultiHeads model as well as four state-of-the-art methods, namely SST~\citep{SST_2022_AAAI}, SARB~\citep{SARB_2022_AAAI}, HST~\citep{HST_2024_IJCV}, and PSScreen V1~\citep{PSScreen_2025_BMVC}. Different to retinal disease screening, the text encoder of semantic decoupling module directly encodes disease names rather than expert descriptions as BioClinicalBERT~\citep{BioClinicalBERT_2019} is pre-trained on large-scale chest X-ray reports and has strong representation capability with chest diseases.

As shown in Tab.~\ref{tab:psl_cxr}, \textbf{PSScreen V2} achieves the best performance on the two multi-label classification datasets CheXpert~\citep{Chexpert_2019_AAAI} and ChestX-ray14~\citep{chestXray14_2017_CVPR}, with improvements of 1.7\% and 2.3\% in $mF$ over the second-best methods PSScreen V1~\citep{PSScreen_2025_BMVC} and SARB~\citep{SARB_2022_AAAI}, respectively. This demonstrates that our \textbf{PSScreen V2} is also applicable to other disease classification, achieving superior performances.

 \begin{table*}[t!]
\centering
\large
\caption{F-score (\%) comparison of different partially supervised learning methods across three chest X-ray datasets. The best and second-best are highlighted in bold and with an underline.}
\resizebox{1.0\textwidth}{!}{
\begin{tabular}{lccccccccccccccccccc}
\toprule
Methods
& \multicolumn{9}{c}{CheXpert}
& \multicolumn{9}{c}{ChestX-ray14} 
& SIIM-ACR \\
\cmidrule(lr){2-10}  \cmidrule(lr){11-19} \cmidrule(lr){20-20}

& Norm & Card & Edem & Cons & Pneu1 & Atel & Pneu2 & Effu & $mF$ 
& Norm & Card & Edem & Cons & Pneu1 & Atel & Pneu2 & Effu & $mF$ &$mF$  \\
\midrule
MultiHeads   & 38.3 & \underline{79.7} & 61.4 & 51.0 & 22.9 & 22.4 & 53.5 & 65.7 & 49.4 & \textbf{41.0} & 35.0 & 37.8 & 43.0 & 23.6 & 17.2 & 62.4 & 53.3 & 39.2  & 70.9  \\
SST        & \textbf{45.6} & \textbf{80.7} & 61.6 & 52.9 & 29.1 & 21.1 & \textbf{61.2} & 67.0 & 52.4 & 38.0 & \textbf{40.4} & \underline{40.8} & 42.1 & 22.3 & 11.3 & \underline{64.5} & 52.3 & 39.0  & 68.3  \\
SARB        & \textbf{45.6} & 76.6 & 60.7 & 56.5 & 31.1 & 22.0 & \underline{56.1} & \underline{70.8} & 52.4 & 38.0 & 35.7 & 39.2 & \textbf{45.9} & \underline{27.7} & 20.8 & \textbf{65.5} & 53.5 & \underline{40.8} & 75.7  \\
HST       & \textbf{45.6} & 76.3 & 58.9 & \textbf{58.3} & \textbf{35.0} & 21.1 & \textbf{61.2} & 70.1 & 53.3 & 38.0 & 34.7 & 39.4 & \underline{45.0} & 27.4 & 11.4 & \textbf{65.5} & \underline{54.8} & 39.5 & 68.2  \\
PSScreen V1 & \underline{45.1} & 78.2 & \underline{64.3} & 53.7 & 31.8 & \underline{39.7} & 54.3 & 70.5 & \underline{54.7} & \underline{38.2} & 35.1 & 37.9 & 41.3 & 24.4 & \underline{25.1} & 64.2 & 53.3 & 40.0 & \textbf{76.5}  \\
PSScreen V2         & 44.8 & 78.7 & \textbf{68.7} & \underline{57.0} & \underline{34.7} & \textbf{43.6} & 52.2 & \textbf{71.7} & \textbf{56.4} & \underline{38.2} & \underline{38.0} & \textbf{42.8} & \textbf{45.9} & \textbf{29.3} & \textbf{30.3} & 61.7 & \textbf{58.2} & \textbf{43.1} & \underline{75.9}  \\
\botrule
\end{tabular}
}
\label{tab:psl_cxr}
\end{table*}

\section{Conclusions}
In this paper, we introduce \textbf{PSScreen V2}, an advanced partially supervised framework for multiple retinal disease screening. By integrating a teacher–student self-training paradigm with two newly designed feature augmentation modules, LF-Dropout and LF-Uncert, \textbf{PSScreen V2} effectively addresses the challenges of partial label absence and domain shift. Experimental results on a wide range of in-domain and out-of-domain fundus datasets show that our method not only surpasses existing SOTAs but also demonstrates strong generalization to unseen domains. In addition, its compatibility with different backbones and successful extension to chest X-ray datasets further validate the universality and adaptability of the framework. Overall, \textbf{PSScreen V2} provides a promising direction for scalable, cost-effective, and generalizable medical image screening in real-world clinical scenarios.

\section*{Acknowledgments} This work was supported in part by the Academy of Finland under Research Fellow Grant No. 355095, and by the Hunan Provincial Natural Science Foundation of China under Grant No. 2023JJ30699. The authors also acknowledge the CSC–IT Center for Science, Finland, for computational resources.

\bibliography{sn-bibliography}

\end{document}